\documentclass[journal]{IEEEtran}
\usepackage{amsmath,amsfonts}
\usepackage{amsthm}
\usepackage[linesnumbered,ruled,vlined]{algorithm2e} % vlined参数添加竖线
\usepackage{array}
\usepackage[caption=false,font=normalsize,labelfont=sf,textfont=sf]{subfig}
\usepackage{textcomp}
\usepackage{stfloats}
\usepackage{url}
\usepackage{verbatim}
\usepackage{graphicx}
\usepackage{cite}
\usepackage{hyperref}
\usepackage{marvosym}
\usepackage[noend]{algorithmic} % 添加 noend 选项
\newtheorem{definition}{\textbf{Definition}}
\newtheorem{example}{\textbf{Example}}
\newtheorem{theorem}{\textbf{Theorem}}
\newtheorem{proposition}{\textbf{Proposition}}
\newtheorem{problem}{\textbf{Problem}}	
\usepackage{multirow}
\usepackage[table,xcdraw]{xcolor}
\usepackage{tabularx}
\usepackage{booktabs}
\usepackage{adjustbox}
\usepackage{color}
\usepackage{pifont}
\definecolor{darkgreen}{rgb}{0.0, 0.5, 0.0}
\definecolor{darkred}{rgb}{0.5, 0.0, 0.0}
\usepackage{enumitem}
\usepackage{pifont}

\usepackage[normalem]{ulem}
\useunder{\uline}{\ul}{}

\begin{document}

\title{Exploring the Heterogeneity of Tabular Data: \\A Diversity-aware Data Generator via LLMs}

\author{
\IEEEauthorblockN{Yafeng Tang, Xiaoou Ding, Jianzhuo Du, Zishuo Yan, Zhuang Ma, Zheng Liang, Zekai Qian, Hongzhi Wang\textsuperscript{\Letter}\thanks{\textsuperscript{\Letter} Hongzhi Wang is the corresponding author.}}
    \IEEEauthorblockA{School of Computer Science, Harbin Institute of Technology, Harbin, China, \\\href{mailto: yangyf@stu.hit.edu.cn}{tangyf@stu.hit.edu.cn},
    \href{mailto: dxo@hit.edu.cn}{dingxiaoou@hit.edu.cn}
    \{\href{mailto: 2023111073@stu.hit.edu.cn}{Dujz},\href{mailto: 2023111352@stu.hit.edu.cn}{yanzs},\href{mailto: MZ@stu.hit.edu.cn}{mazhuang}\}@stu.hit.edu.cn,
    \{\href{mailto: liangzheng980224}{liangzheng},\href{mailto:qzk010728@gmail.com}{qianzekai}\}@gmail.com,
    % \href{mailto: hit_wyc@proton.me}{wangyc@proton.me},
    \href{mailto: wangzh@hit.edu.cn}{wangzh@hit.edu.cn}
    }
}

\markboth{Journal of \LaTeX\ Class Files,~Vol.~XX, No.~X, August~2025}%
{Shell \MakeLowercase{\textit{et al.}}: Exploring the Heterogeneity of Tabular Data: \\A Diversity-aware Data Generator via LLMs}


\maketitle
\footnote{Submitted to IEEE Transactions on Knowledge and Data Engineering (TKDE) for peer review.}
\begin{abstract}
Tabular data generation has become increasingly essential for enabling robust machine learning applications, which require large-scale, high-quality data. Existing solutions leverage generative models to learn original data distributions. However, real-world data are naturally heterogeneous with diverse distributions, making it challenging to obtain a universally good model for diverse data generation.
To address this limitation, we introduce \textbf{D}iversity-\textbf{A}ware \textbf{T}abular data g\textbf{E}nerator (DATE), a framework that
(i) prepares high-quality and distributionally distinct examples for in-context learning by effectively partitioning the original heterogeneous data into multiple diverse subsets; (ii) harnesses Large Language Models (LLMs) to explore the diversity of the partitioned distribution with decision tree reasoning as feedback, generating high-quality labeled data for each subset. 
However, the massive generated data inherently involves a trade-off between diversity and quality. To integrate this issue, existing solutions greedily select ``validation-best'' data. However, we prove that the selection in heterogeneous settings does not possess the greedy-choice property, and design a Multi-Arm Bandit-based sampling algorithm that balances the diversity and quality of generated data.
Extensive experiments on tabular classification and regression benchmarks demonstrate that DATE consistently outperforms state-of-the-art GAN-based and LLM-based methods. On average, DATE achieves a 23.75\% reduction in error rate with just 100 generated data. Empirically, we demonstrate that data generated by DATE can improve the accuracy of Direct Preference Optimization (DPO) and enhance the reasoning capability of LLMs on the target data. Code is available at \href{https://github.com/windblow32/DATE}{https://github.com/windblow32/DATE}.

% % (ii) harness Large Language Models (LLMs) to discover the cause of diversity and generate customized labeled data for each subset. More importantly, to enhance the fitness of synthetic data, we design a data-driven sampling algorithm that selects diversity-relevant synthetic subsets for on-demand augmentation.
% Extensive experiments on various classification benchmarks demonstrate that our approach consistently outperforms state-of-the-art methods by adding only several data tpoints o each partitioned subset. (achieving up to xxx absolute improvement in error rate) In particular, with the generated diversity data, our solution can address the overfitting problem posed by long-tail distributions.
\end{abstract}


\vspace{-1em}
\section{Introduction}
% 针对大语言模型，充足的数据量与数据的多样性均必不可少
% 引出表格数据生成，讲清由于隐私，标注成本等原因，不依赖外部数据的生成式方法是主流
\IEEEPARstart{T}{abular} data is the core representation in database systems. The reliability of these systems, especially when supporting machine learning (ML) workloads, is built on the high-quality and large-scale tabular datasets. However, issues such as data scarcity, heterogeneous distributions~\cite{heterogeneous}, making ML models unreliable and fail to generalize.
For example, MIT researchers have shown that although hundreds of AI tools were created for COVID-19 detection, none succeeded in practice with the heterogeneous clinical histories.
% particularly for rare yet crucial events, often significantly degrade the performance of machine learning (ML) models. 
To make matters worse, assembling high-quality datasets is often costly and infeasible due to annotation burden.
% and often infeasible due to annotation burden and privacy concerns. 
As a result, synthetic tabular data generation~\cite{shi2025comprehensivesurveysynthetictabular} has gained attention as a compelling alternative to mitigate data scarcity and strengthen downstream model robustness.

% 介绍主流数据生成方法，分为基于非语言模型和语言模型两类
\textbf{Tabular data generation solutions.}
Synthetic tabular data generation has been an open problem for decades, which is often built upon generative models. Conventional solutions are typically built on non-language models, such as VAEs~\cite{2014PrivBayes}, diffusion models~\cite{RelDDPM, CoDi}, and GANs~\cite{table-GAN, PATE-GAN, ITS-GAN, CTGAN} to reproduce joint feature–label distributions. As the rise of Large Language Models (LLMs)~\cite{TABSYN}, fine-tuning
% Recently, Large Language Models (LLMs) have emerged as promising alternatives~\cite{TABSYN}. F
adapts LLMs to domain-specific patterns with task-specific examples~\cite{GReaT, HARMONIC}, and prompt-based generation exploits in-context learning to reduce reliance on training data~\cite{EPIC, CLLM}. In summary, all these methods rely on learning distributions from examples to generate identically distributed data. 

\begin{figure}[t]
    \centering
    \includegraphics[width=\columnwidth]{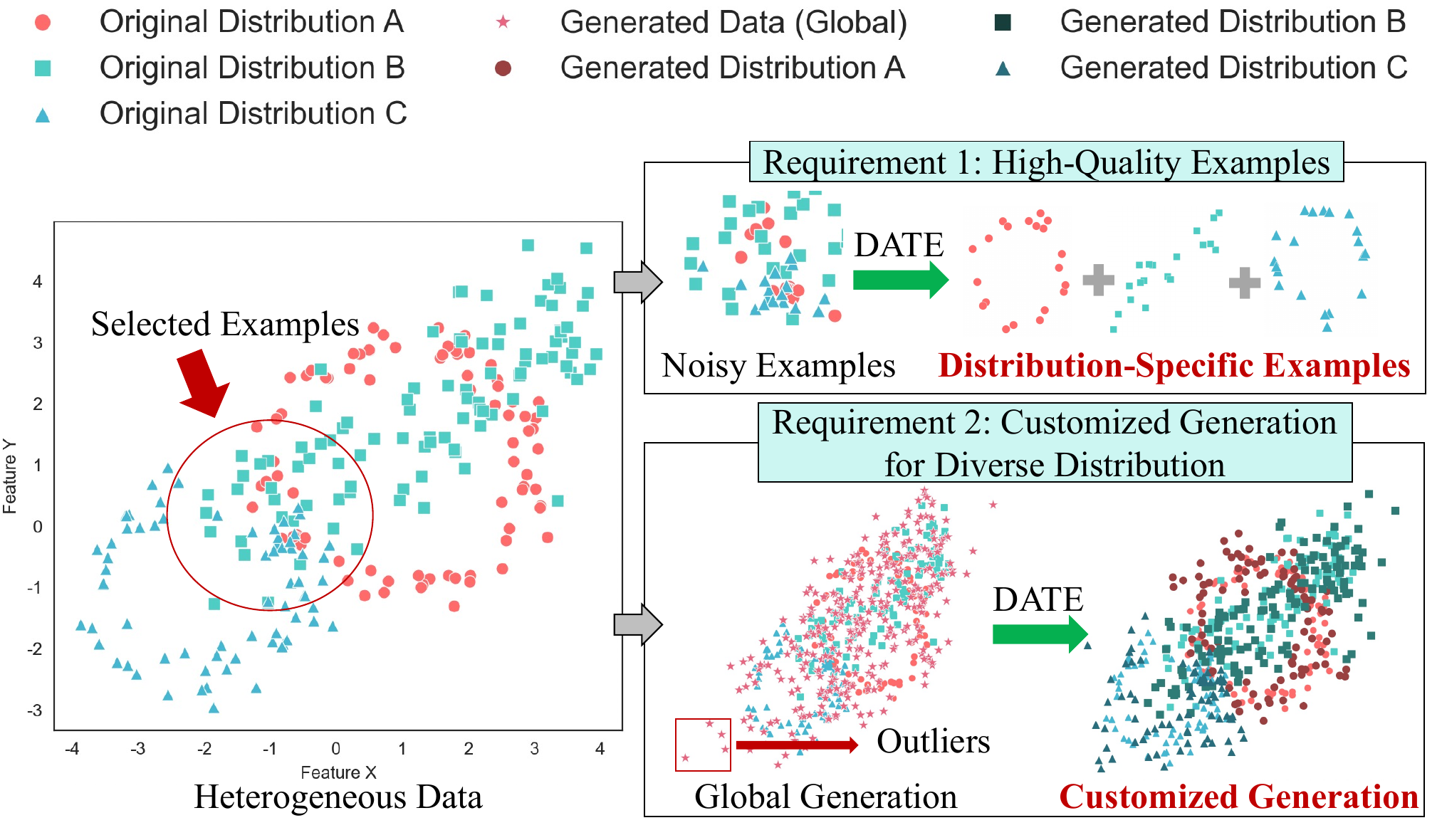}
    \caption{Two key requirements for heterogeneous data generation (better viewed in color): \ding{172}\textbf{High-Quality Examples} and \ding{173}\textbf{Customized Generation for Diverse Distribution}. We present a toy example composed of three distributions.}
    \label{fig: intro}
    \vspace{-2em}
\end{figure}

\textbf{Requirements from Heterogeneous Tabular Data.}
Real-world tabular datasets are naturally heterogeneous~\cite{heterogeneous}, such as disease diagnosis based on clinical histories~\cite{hospital}, IEEE-CIS~\cite{fraud} for fraud detection that integrates transaction records from different regions, and HadISD~\cite{weather} for weather forecasting that aggregates observations from various climate zones. As shown in Figure~\ref{fig: intro}, this inherent heterogeneity proposes the following requirements for the real-world applications of existing data generation solutions.
% 更紧凑的版本
\begin{enumerate}[label=\ding{172}, nosep, leftmargin=*]
\item \textbf{High-Quality Examples.} In heterogeneous settings, data can be viewed as a mixture of multiple distributions. Thus, if the selected examples are too noisy to represent the global distribution,
% In heterogeneous settings, data from each distinct distribution can be regarded as a subset. If selected examples for a subset are noisy, 
generators may hallucinate spurious patterns and produce meaningless outliers. Therefore, a practical solution needs a high-quality set of examples that represent the key dependencies of the whole dataset.
\end{enumerate}

\begin{enumerate}[label=\ding{173}, nosep, leftmargin=*]
\item \textbf{Customized Generation for Diverse Distribution.} A single model often struggles to capture the heterogeneous distributions in real-world tabular data. While structured approaches such as ensemble learning and mixture-of-experts-based LLMs~\cite{ashiga2025ensemblelearninglargelanguage} offer promising mechanisms for modeling such distributional diversity, their potential is largely underexplored in tabular data generation. To bridge this gap, an effective data generator needs to be capable of inferring the characteristics of diverse distributions and generating data aligned with each specific distribution.
\end{enumerate}

% 结合上面的需求，讲清楚每种方法的缺点，引出我们需要使用决策树+子集划分搞出多样性+解释推理,同时讲清楚finetuning为什么不如prompt，引出后面的challenge中针对prompt-based的LLM生成，具体的两个挑战
\textbf{Limitations of existing solutions.}
To the best of our knowledge, few synthetic tabular data generation solutions can address the above two issues simultaneously. Non-language model-based solutions like Table-GAN~\cite{table-GAN} often suffer from low-quality examples and the model collapse problem. State-of-the-art LLM-based methods such as GReaT~\cite{GReaT} and EPIC~\cite{EPIC} also require high-quality prompts for fine-tuning or iterative prompt engineering to fit the heterogeneous distribution directly. However, the quality of synthetic data was limited by a learned yet suboptimal distribution. To meet the demands mentioned above, a natural idea is to partition heterogeneous data into high-quality and distributionally coherent subsets, enabling customized prompt designing and generation. However, fine-tuning LLMs still require a higher cost to train many distribution-specific models, while prompt-based methods~\cite{NIPS24} leverage in-context learning with a single foundation model that can dynamically adapt to diverse distributions through conditional prompting. In this paper, we focus on prompt-based solutions, which have been generally adopted~\cite{NIPS24, EPIC, CLLM} for reducing dependence on extensive reference data and avoiding overfitting when reference data are limited.

% As the new favorite of tabular data preparation, we observe that tree-based solutions have demonstrated great performance among many tasks, including data generation~\cite{}. Besides, the abilities of providing rule-based reasoning information and explainable feedback are suitable for LLM. Motivated by these observations, it is promising to leverage tree-based models to prepare high-quality and distribution-customized data for LLMs data generation.

\subsection{Challenges}
Prompt-based LLM methods face two challenges caused by heterogeneous tabular data as follows:
% \lz{'Prompt Designing' and 'Distribution-Specific Generation' don't seem challenging. Also 'Diverse Distributions' and 'Heterogeneous Data' seems too similar' that the reader might doubt the necessity of two challenges(instead of one). Consider make them orthogonal.}

%两点挑战：（1）prompt需要便于LLM 理解异构数据并reasoning，涉及子集划分，构建in-context learning example （2）生成过程要学习不同分布间特性，追溯多样性成因，并且While synthetics can enhance individual subsets,their combination may degrade global performance. Identifying distribution-qualified samples from the numerous generated data remains the pivotal challenge in heterogeneous tabular data generation.

\textbf{Challenge 1: High Costs in Constructing High-Quality Examples.}
% Prompt Designing for Diverse Distributions.}
Heterogeneous data are naturally composed of diverse distributions. We can not expect that a single set of chosen examples for in-context learning will guide LLM to perform well across all distributions.
Existing solutions often fail to provide high-quality example sets for each distribution within the heterogeneous data.
Moreover, when selecting the high-quality data for each distribution, we actually need to consider every combination of records as the candidate subset. In this case, the candidate space is $O(2^n)$, and searching for each candidate requires a partitioning model retraining. As a result, constructing such examples is really time-consuming.

\textbf{Challenge 2: Low Fidelity in Distribution-Specific Generation.}
Prompt-based methods rely on the inherent knowledge during pre-training, which often causes LLMs to overlook the specific distributional patterns intended by the carefully designed prompts, leading to overfitting. What's worse, in real-world scenarios, heterogeneous data often lacks explicit dependency structures or column descriptions, thereby providing insufficient contextual guidance. 
% Therefore, existing methods struggle to generate sufficiently tailored and diverse data that adequately capture the distinct distributions within heterogeneous data. 
Even if diverse data can be generated for heterogeneous distributions, effectively integrating them remains challenging. The common greedy selection of a ``validation-best'' subset often overfits and fails to generalize, highlighting a key challenge in heterogeneous tabular data synthesis.
% Even if we generate sufficiently tailored and diverse data that adequately capture the distinct distributions within heterogeneous data, integrating them remains nontrivial.
% A common practice for most solutions by greedily selecting the ``validation-best'' subset may achieve high validation scores, yet perform poorly on the test set. 
% This issue remains a challenge in heterogeneous tabular data generation.

% Furthermore, an inherent trade-off emerges between diversity and quality during the post-generation filtering of synthetic data. It is noteworthy that a common practice of greedily selecting the ``validation-best'' subset may achieve high validation scores, yet perform poorly on the test set. This issue remains a challenge in heterogeneous tabular data generation.

%%%%%%建议在此处深入讨论一下当前方法的问题和我们解决问题的思路，引出本文研究问题的方法论，几句话即可

\subsection{Contributions}
In this paper, we conduct an in-depth exploration of the data generation problem in the heterogeneous tabular data setting. Existing methods often overlook this inherent heterogeneity, thus struggling with low-quality examples and suboptimal distributions. To address this, we propose a novel prompt-based method by initializing rules
% \lz{better give an example for such rules here} 
for constructing high-quality prompts, leveraging LLMs' reasoning to refine these rules automatically, and then using the refined rules to drive the generation process. Our major contributions in this paper are summarized as follows:

(1) We present the Diverse-aware Tabular data generation (DATE) solution, which harnesses LLM to explore the diverse distributions within heterogeneous data for customized generation. 
To bridge the semantic gap between heterogeneous data and prompts for better LLM comprehending, we design the Distribution-Guiding Rule (DGR) in Section~\ref{sec:Preliminary}, which provides interpretable guidance in a controllable manner.
% We design the Distribution-Guiding Rule (DGR) as a bridge connecting real-world data distributions with prompt construction.\lz{Why use a rule to do this bridging? What are the advantages of Rules?}

%%每段第一句呼应challenge内容，后面which引出如何解决的
(2) To prepare examples from heterogeneous data for in-context learning, we propose the DGR Discovery algorithm in Section~\ref{sec: DGR-based Prompt Designing}, which can efficiently partition heterogeneous data into high-quality distributions with less model retraining, and construct them into examples for customized data generation.
% construct them into high-quality examples for in-context learning.\lz{What is the theoretical or conceptual advantages of this Algorithm(compared to brute-force)? Maybe time/space complexity or some theoretical guarantee?}

(3) To achieve distribution-specific LLM reasoning, we design the DGR-guided generation algorithm in Section~\ref{sec: DGR-Guided Iterative Data Generator}, which iteratively refines DGRs to capture the unique dependency within each distribution, thereby mitigating overfitting.
To adaptively filter synthetic data toward an optimal diversity-quality balance, we develop a sampling algorithm with a controllable error bound in Section~\ref{sec: MAB}.
% The core principles of this algorithm are also applicable to other tasks with the same optimization objective, such as important data selection and iterative data generation.}
% After that, we develop a sampling algorithm\lz{What is the motivation for sampling?} to adaptively filter synthetic data while maintaining diversity-quality balance in Section~\ref{sec: MAB}, which can also apply to important data selection methods and other iterative data generation tasks.\lz{'important data selection' and 'iterative generation' also seem strange and informal? And why do these two problems matter?}

(4) In Section~\ref{sec:experiemnt}, we evaluate DATE on 8 classification and 2 regression tasks, which are highly practical and relevant to critical domains such as healthcare and finance. Compared to GAN-based, prompt-based, and fine-tuning-based solutions, DATE achieved an average reduction of 14.3\% in error rate on classification tasks and 61.5\% in MSE on regression tasks with 89\% less generated data.
% \lz{Why is less generated data good? Does it mean lower computational and token cost? Or lower the ratio of synthesis data in real-world datasets? Justification is need but maybe not here}. 
% Ablation studies further validate that \textcolor{blue}{DATE-generated data} can improve the accuracy of the direct preference optimization algorithm \textcolor{blue}{on classification tasks} by an average of 11.0\%, demonstrating \textcolor{blue}{its effectiveness} in enhancing the reasoning capabilities of LLMs.

\section{Overview}
In this section, we introduce the background for tabular data generation (Section~\ref{sec:Background}), elaborate on the key components of our prompt (Section~\ref{sec:Preliminary}), present the essential problems in DATE (Section~\ref{sec:Problem}), and give an overview of our solution (Section~\ref{sec:Framework}).

\subsection{Background}\label{sec:Background}
Tabular Data Generation(TDG) solutions consist of at least two phases: column-level feature synthesis and row-level record generation~\cite{TDA}. Considering the heterogeneous nature of tabular data~\cite{heterogeneous}, one significant challenge in ML is the uneven distribution of samples, particularly in long-tailed data~\cite{EPIC}. In this paper, we focus on row-level record generation to obtain more training samples, thereby increasing their categorical variety and altering the data distribution. We treat column-level feature synthesis as an orthogonal problem.

\begin{definition}[Tabular Data Generation]
    Given an original table $T^{O}$ with the attribute set $\mathcal{A}$ and the record set $\mathcal{R}$, the augmented output of Row-Level TDG $T^{A}$ satisfies:
    % \lz{I see your modification. But I still feel that the problem of this paper has not been formally defined, which is still a drawback.}
\begin{equation}\nonumber
    (T^{O}.\mathcal{A}=T^{A}.\mathcal{A}) \wedge (T^{O}.\mathcal{R} \subset T^{A}.\mathcal{R})
\end{equation}
\end{definition}

Generally, TDG aims to generate $T^{A}$ such that a model trained on it achieves strong performance on the test set.
While large language models (LLMs) now achieve state-of-the-art results in this task~\cite{EPIC, CLLM, TABSYN}, they often underperform on real-world heterogeneous data due to a lack of high-quality and distribution-specific samples for fine-tuning. Prompt-based methods address this issue by using in-context learning to guide generation. Thus, effectively handling heterogeneous data reduces to carefully designing high-quality prompts for distribution-specific generation.
% struggle with real-world heterogeneous data, where insufficient distribution-specific data leads to suboptimal fine-tuning. 
% Fortunately, prompt-based methods address this by using in-context learning to steer data generation through carefully designed prompts. Therefore, the core of handling heterogeneous data hinges on designing distribution-specific prompts for diverse data generation.\lz{Is this paragraph written by AI? Be natural}

% Recently, large language models (LLMs) have achieved SOTA performance for generating tabular data~\cite{EPIC, CLLM, TABSYN}. However, for real-world heterogeneous data with diverse distributions, it is hard for these models to find enough distribution-specific data for finetuning, resulting in suboptimal performance. Fortunately, prompt-based approaches leverage in-context learning to impact the distribution of generated data by carefully designing the optimal prompt. Therefore, the core of these solutions for handling heterogeneous data is to design distribution-specific prompts for diverse data generation. 
% In this paper, we explore the prompt-based LLM solver in the heterogeneous data setting. 

\textbf{Motivation of DGR.}  The key to prompt designing is to enable LLMs to comprehend heterogeneous data. 
Recently, the TreePO~\cite{TREEPO}, a high-scoring paper at ICLR 2026 from ByteDance's team, demonstrates that organizing the LLM reasoning path into a tree-like structure—where different answers often share the same initial reasoning steps before diverging later—can significantly improve training efficiency and accuracy.
% Recent research shows that LLM reasoning path naturally exhibits a tree-like structure~\cite{TREEPO}\lz{Where was this paper published? This is a important claim and should have a strong reference.}, where different answers often share the same initial reasoning steps before diverging later. 
Motivated by this, we propose the \textbf{Distribution-Guiding Rules (DGR)}, a tree-structured rule that can describe diverse distributions in heterogeneous data for prompt designing. Next, we will introduce DGR in detail.
% \lz{The `tree' thing should be verified later with example figures, theoretical analysis, or evaluation.}
% which partitions the original data into tree-structured rules and structures it into examples for in-context learning.  Next, we will introduce how we design DGR to support the representation of diverse distributions.

\subsection{Preliminaries}\label{sec:Preliminary}
% Our key idea is to leverage the reasoning capabilities of LLMs to understand and identify effective generation rules without manually specifying the search space, and provide language-based reasoning information highlighting past experiments as feedback for iterative rule improvements. Therefore, in this section, we will introduce how we design \textit{subset generation rules}, which is the core of our prompt to make LLM better understand heterogeneous data. \textcolor{red}{For detailed prompts, please refer to our code at https://xxx.com}
% In this section, we will introduce \textit{Distribution-Guiding Rules (DGR)}, which is the core of our prompt to make LLMs better reason heterogeneous data\lz{`better reason heterogeneous data' is not a good expression.}.
In this section, we introduce Distribution-Guiding Rules (DGR) as the core component of our prompt for enhanced LLM reasoning over heterogeneous data.

We start with the formal definition of each component of DGR, including the predicate space $P$ and the conditions $C$. Then, we give a formal definition of DGR.

We briefly introduce the predicate space as the foundation of DGR, so that a conjunction of predicates could describe a simple set of data for a specific distribution~\cite{crr}.
\begin{definition}[Predicate Space]\label{def: Predicate}
Given a heterogeneous table $T$ with the attribute set $\mathcal{A} = \{A_1, A_2, \dots, A_m\}$ and the operator set $\Phi = \{>, \geq, <, \leq\}$, the predicate space $P$ is:
\begin{equation*}
P = \{ p=A \phi c | A\in T.\mathcal{A}, \phi\in \Phi \}
\end{equation*}
where $c$ is a constant from the domain of the attribute $A$.
\end{definition}

% Given tuple $t\in T$, assume that $t$ is satisfied by any built-in predicates, we have $t \models p$ if $t.A\phi c$. 
Based on the predicate space $P$, we can describe a simple part of the data by $\bigwedge_{p\in pred(r_{i})}p$. However, in heterogeneous data, data following the same distribution can be scattered across multiple parts. To capture this unified distribution, we propose \textbf{Distribution-Guiding Rules (DGR)} as follows.
% Based on the predicate space $P$, we introduce conditions $C$ to describe a simple distribution of data as follows:
% \begin{definition}[Distribution-Guiding Rule]
%     Given the predicate space $P$, the distribution-guiding rule $r$ is a conjunction composed of a set of predicates $pred(r_{i}) \subseteq P$
%     \begin{equation*}
%         r_{i}: \bigwedge_{p\in pred(r_{i})}p
%     \end{equation*}
% \end{definition}

% \begin{definition}[Condition]
%     Given the predicate space $P$, a conjunction $C_{i}$ is composed of a set of predicates $pred(C_{i}) \subseteq P$: %with the conjunction
%     \begin{equation*}
%         C_{i}: \bigwedge_{p\in pred(C_{i})}p
%     \end{equation*}
% \end{definition}
\begin{definition}[Distribution-Guiding Rule]\label{def: DGR}
    Given a predict space $P$, conditions $C_{i}$, the Distribution-Guiding Rule is a DNF composed of countable conjunctions $r_{1}... r_{n}$:
    \begin{equation*}
        r:(\bigwedge_{p\in pred(r_{1})}p)\vee...\vee(\bigwedge_{p\in pred(r_{n})}p)
    \end{equation*}
\end{definition}

Data sharing the same DGR can be grouped into a subset. The subset and DGR are then structured as a rule-data example in prompts (see Section~\ref{sec: DGR-based Prompt Designing} for details) for in-context learning.

%%%这一段是不是谈一旦咱们问题定义的优点
\textbf{Advantage.} DGRs adopt the interpretable form of decision tree rules~\cite{breiman1984classification}, steering LLMs to comprehend diverse distributions in heterogeneous data. Moreover, by controlling the data quality of partitions defined by DGRs, we can assess the quality of DGRs themselves, thereby guiding LLMs to generate high-quality data. We formalize the efficient and quality-aware DGR searching as Problem~\ref{problem: DGR Searching} below.

% Note that although Distribution-Guiding Rules follow the same form of intuitive Decision Tree node~\cite{breiman1984classification}, our target is completely different from single model training, which will be introduced as Problem~\ref{problem: DGR Searching} in Section~\ref{sec:Problem}.

\subsection{Problem Definition}\label{sec:Problem}
In this section, we consider the problem of prompt-based generation in heterogeneous data scenarios, including two parts: prompt designing and data generation. We first introduce the DGR searching as Problem~\ref{problem: DGR Searching} to partition heterogeneous data into high-quality subsets. Based on the prompt formed by DGRs and subsets, we formalize the distribution-specific data generation as Problem~\ref{problem: DSDG}.

For the prompt designing part, due to the vast space formed by the constant $c$ and the attribute set $\mathcal{A}$ in the predicate space $P$, manually crafting DGR is intractable. Therefore, we first focus on DGR searching as follows:
\begin{problem}[Distribution-Guiding Rules Searching]\label{problem: DGR Searching}
    Given a table $T$ with attribute set $\mathcal{A} = \{A_1, \dots, A_m\}$, an evaluation model $m$, and an error tolerance threshold $\rho \in [0,1]$, find the most representative rule set $R$ satisfying:
    \begin{gather}\nonumber
        minimize \quad |R| \\
        R = \{r | r : \bigvee_{j=1}^{n}(\bigwedge_{p \in pred(r_{j})}) \}, error(M, T_{r})<\rho \nonumber
    \end{gather}
    where
    \begin{gather}\nonumber
        T_r = \{ t \in T \mid t \models r \} \\
        error(M,T_{r})=\begin{cases}
            1-\frac{|\{t\in T_{r} \mid m(t)=y_t\}|}{|T_{r}|} & \text{(classification)} \\
            \frac{1}{|T_{r}|}\sum_{t\in T_{r}}|m(t)-y_t| & \text{(regression)}
        \end{cases} \nonumber
    \end{gather}
\end{problem}

A brute-force traversal in the search space of DGR would acquire $O\left(\sum_{A_i\in {A}}|\mathrm{domain}(A_i)|\times|\Phi|^d\right)$ 
, where $d$ is the max depth of DGR. In Section~\ref{sec: DGR Discovery}, we propose an $O(|T|^2log|T|)$ algorithm, which effectively decouples complexity from the domain size of attributes and makes it scale solely with the dataset cardinality. Afterwards, the $(T_r,r)$ pairs are integrated into in-context examples. Building upon the examples, we formalize the distribution-specific data generation task as follows:
\begin{problem}[Distribution-Specific Data Generation]\label{problem: DSDG}
    Given an original table $T^{O}$ with the attribute set $\mathcal{A}$, the DGR set $R = \{R_{1},...,R_k\}$, Distribution-Specific Data Generation aims to generate the augmented table $T^{A}$ satisfying:
    \begin{gather}\nonumber
        T^A = \bigcup_{r \in {R}} LLM(T_r^O,r) \\
        \forall r \in R, (T^O_r \subseteq T^A_r)\wedge (T^O_r.\mathcal{A} = T^A_r.\mathcal{A}) \nonumber
    \end{gather}
    where each DGR $r\in R$ represents a specific distribution. 
\end{problem}

To solve Problem~\ref{problem: DSDG}, DGR guides LLM to enrich the augmented table $T_r^A$ with numerous records, which inherently navigates a trade-off between quality and diversity.
However, for heterogeneous data, we demonstrate that simply selecting the best-performing results on a validation set fails to achieve global optimality in Section~\ref{sec: MAB}. Consequently, sampling generated data from LLM outputs remains a challenging task.

\begin{figure*}[htbp]
    \centerline{\includegraphics[width=\textwidth]{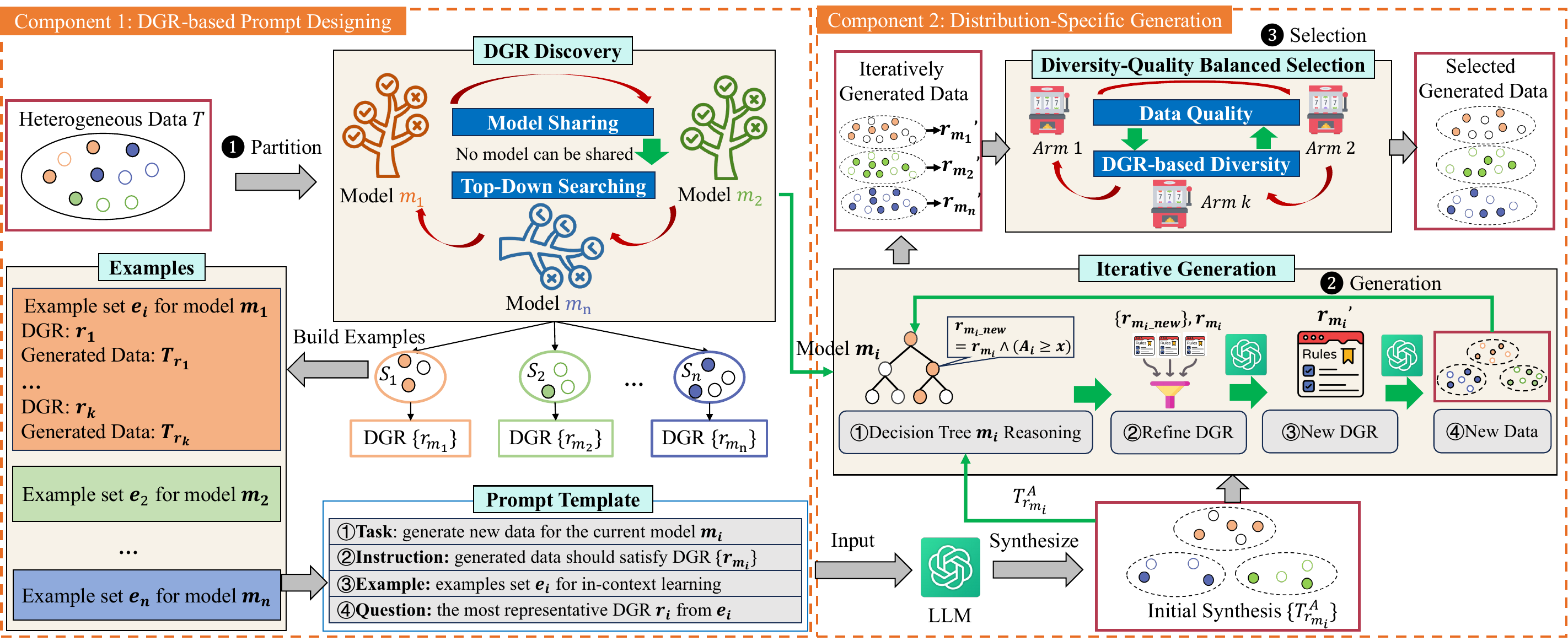}}
    \caption{Overall framework of DATE (better viewed in color), which consists of two components with three \textbf{steps}: (1) DGR-based Prompt Designing, which \textbf{partitions} heterogeneous data into diverse data distributions through DGR discovery and constructs prompts with DGR-based examples. We present the prompt template at the bottom center. (2) Distribution-Specific Generation, which focuses on a targeted shared model $m_i$
    and iteratively \textbf{generates} data with refined DGRs. To integrate generated data, DATE \textbf{selects} diversity-quality balanced synthesized data as the final result.}
    \label{fig:DSDG workflow}
\end{figure*}

\subsection{Solution Overview}\label{sec:Framework}
In this section, we introduce two main components of our DATE framework. 

\textbf{Motivation.} 
% Our key idea is to initialize DGR for building high-quality and distribution-specific prompts, and then leverage the reasoning capabilities of LLMs to refine DGR without manually specifying the search space, and provide language-based reasoning information highlighting past experiments as feedback for iterative rule improvements.
Our key idea is to initialize DGRs for high-quality prompts, and then iteratively refine them based on language-based feedback from past LLM experiments, thereby automating the search without manual specification.
Finally, we can employ refined DGRs for data generation. 

\textbf{DATE Workflow.} DATE consists of two components as shown in Figure~\ref{fig:DSDG workflow}. DATE first deploys the DGR-based Prompt Designing Module to partition heterogeneous data into diverse subsets and construct them into high-quality examples for prompt engineering. Subsequently, the Distribution-Specific Generation Module extrapolates DGRs for each subset to iteratively generate diverse data, while incorporating a sampling algorithm 
%that strategically balances 
to balance the diversity-quality tradeoff in final outputs. The details are as follows.

\textbf{DGR-based Prompt Designing Module. }
To enable LLM reasoning on heterogeneous data, a thorny task is to design high-quality prompts for diverse distributions. Therefore, the DGR Discovery algorithm partitions heterogeneous data with DGRs, ensuring $error(m, T_r)<\rho $ within each derived subset $T_r$ for a given error threshold $\rho$. We accelerate the DGR discovery algorithm through the model sharing and the top-down searching strategies. The template of the DGR-based prompt is presented at the bottom of Figure~\ref{fig:DSDG workflow}.

\textbf{Distribution-Specific Generation Module.}
DATE generates the initial dataset with the most representative DGR sorted by DGR discovery algorithm in the first module.
% The first module generates the initial dataset with the most representative DGR sorted by DGR discovery algorithm.
To further generate more diverse data, our second module optimizes LLM reasoning for the current distribution by inferring new DGRs from decision tree reasoning as interpretable feedback and refining them to guide iterative data generation. Finally, to select the diversity and quality-balanced results from the numerous generated data, we design a multi-armed bandit-based sampling algorithm with a controllable error bound.

\section{DGR-based Prompt Designing}\label{sec: DGR-based Prompt Designing}
% As introduced in challenge 1, the difficulty of designing prompts for heterogeneous tabular data lies not merely in identifying diverse distributions, but in representing them in a form that is both mathematically precise and semantically interpretable for LLMs. 
In this section, we focus on the prompt design in DATE for Challenge 1. The prompt template in Figure~\ref{fig:DSDG workflow} can seamlessly handle data heterogeneity, as DGR clarifies the constraint on each distribution for LLM generation. We first discuss how we use DGR for DATE prompt engineering in Section~\ref{sec: prompt designing}. Then we introduce how to efficiently discover DGR in Section~\ref{sec: DGR Discovery}.

\vspace{-1em}
\subsection{DGR-based Prompt Template}\label{sec: prompt designing}
As the basis of our DATE framework, DGR and DGR-based examples are the core designs in our prompt template shown in Figure~\ref{fig:DSDG workflow}, inspired by the rule-based regression~\cite{crr} and in-context example selection paradigm~\cite{lz}. 
%We will give the formal definition of DGR-based examples, and highlight that the quality of the DGRs determines the quality of DGR-based examples.

\textbf{Limitation of Existing Prompts for Generation. }
In prompt-based data generation, LLMs learn from a small set of labeled and representative examples, known as in-context examples. Formally, a prompt $P$ combines instructions, tasks, questions, and examples $E = \{e_1, \cdots, e_n\}$ to condition LLM outputs $O \sim LLM(P)$. Therefore, the quality of $E$ often determines whether $O$ respects the intended constraints between columns. However, existing methods~\cite{EPIC,NIPS24} primarily focus on providing feedback to LLMs during generation, while overlooking the quality of the data within the prompt itself. 

\textbf{DGR-based Examples for Heterogeneous Tabular Data. }
To address the above limitation, we propose DGR-based examples as follows. In Definition~\ref{def: examples}, each example is anchored by a DGR, which represents a specific distribution within heterogeneous data. By ensuring that the data partitioned by each DGR maintains a controlled error, we guarantee the high quality of the resulting DGR-based examples. The DGR-based examples are sampled from the data subset adhering to a specific distribution via the stratified sampling algorithm~\cite{pmlr-v139-lu21d}. The definition of DGR-based examples is as follows. 

\begin{definition}[DGR-based Examples]\label{def: examples}
    Given DGR set $R$, the table $T$ with attribute set $\mathcal{A}$, an evaluation model $m$, the error threshold $\rho$, the DGR-based example $E$ is:
    \begin{equation*}
        E:(m,\rho,r,T_r)
    \end{equation*}
    where $m: X\xrightarrow{}Y$ is an evaluation model mapping from the value of attributes $X\subset \mathcal{A}$ to the prediction of target $Y\in \mathcal{A}$ with $\rho>0$ as the maximum error between the prediction and the value of $Y$\footnote{The threshold $\rho$ and evaluation model $m$ stems from Algorithm~\ref{alg: partition}, in which a tighter threshold makes DATE identifies the distribution better, so that DGR-based examples lead to better performance.}, and $R$ has no predicates on attribute $Y$. $T_r$ denote the set of all tuples in $T$ that satisfy rule $r$.
    % $T_r = \{ t \in T \mid t \models r \}$
\end{definition}

% \textbf{Components of DGR-based Examples.}\lz{What are these meaning targeted?}
% %The tuple above forms the atomic unit of an in-context example, where 
% In Definition~\ref{def: examples}, $R$ provides a description of the current distribution, $T_R$ contributes representative records\lz{what is representative records}, and $M$ certifies example quality under a fixed tolerance. %$\rho$ ($classification: 1 - accuracy \leq \rho; regression: MSE \leq \rho$). 

\textbf{Importance of DGR Discovery for Prompt Designing.} 
% Based on Definition~\ref{def: examples}, 
Low-quality DGR $r$ would propagate spurious correlations between columns in the partitioned $T_r$, which directly affects LLMs' distributional fidelity~\cite{ADA-ICL}. What's worse, the window of in-context learning has a limited length. Therefore, we next discuss how to discover a concise set of high-quality DGRs.

\subsection{DGR Discovery}\label{sec: DGR Discovery}
In this section, we discuss how to use the Decision Tree to effectively discover concise and high-quality DGRs to construct DGR-based examples.
% \lz{Is this algorithm a variant of CRR? If so, why no reference of CRR? Which parts are your contribution and which parts are contribution of CRR?}

\textbf{Foundation of DGR Discovery.}
A high-quality DGR set can be guaranteed by the error threshold $\rho$, and a concise DGR set is a minimal cover of predicates. In fact, the key to constructing a concise DGR set is to select predicates from $P$, which is shown in the following proposition.

\begin{proposition}[Induction]\label{prop: induction}
    Given the table $T$, an evaluation model $m$, the error threshold $\rho$, DGR $r_1$, if there exists DGR $r_2\vdash r_1$, then $e_1:(m,\rho, r_1,T_{r_1})$ implies $e_2:(m,\rho, r_2,T_{r_2})$.
% \begin{proof}
%     Consider any tuple $t$ with $t\models e_1$. If $t\models e_2$, there exist some conjunction $C_{r_2}\in r_2$ having $t \models C_{r_2}$. Given $r_2 \vdash r_1$, for any $C_{r_2}\in r_2$, there exists conjunction
% $C_{r_1}\in r_1$ s.t. $C_{r_2}\in C_{r_1}$. Thus, for any $C_{r_2}\in r_2$, if $t \models C_{r_2}$, then $\exists C_{r_1} \in r_1, t \models C_{r_1}$. It follows $t \models r_1$. According to $t \models e_1$, we have $error(m,T_{r_1})\leq \rho$, i.e., $t \models e_2$ as well. Otherwise, for $t \not\models r_2$, it naturally has $t\models e_2$. 
% % To sum up, $e_1$ implies $e_2$. 
% \end{proof}
\end{proposition}

Due to limited space, we leave the proofs of all properties and theorems in the Appendix\footnote{\url{https://github.com/windblow32/DATE/appendix.pdf}}.

% Proposition~\ref{prop: induction} guarantees that for any tuple $t$, if $t$ satisfies $e_1$, then $t$ also satisfies $e_2$ under refined $r_2$ with $r_2 \vdash r_1$.  
When discovering DGRs, Proposition~\ref{prop: induction} lays the foundation for the conjunction of the predicates. Next, we will introduce how we construct DGR with these predicates.

\textbf{The Decision Tree Solution.}
In Definition~\ref{def: DGR}, we introduce the search space of DGR, where predictions over parts of data are too exhaustive to explore. 
To take advantage of the subset partitioning methods~\cite{crr} while providing LLM-friendly prompts, we design the DGR discovery algorithm, which naturally extends them to the LLM prompt design setting. DGR shares the same structural form as decision tree nodes, allowing us to directly integrate interpretable tree paths into DGR-based examples as prompt components.
% Fortunately, DGR has a similar structure to the nodes of tree-based models, which can construct the predicate space $P$ for DGR. Moreover, using the decision tree model as the evaluation model $m$ can provide interpretable reasoning paths for the data generated by LLMs. 
However, if a DGR $r$ constructed from decision‑tree predicates induces a subset $T_r$ whose quality falls below the tolerance $\rho$ (i.e., $error(m, T_r) \geq \rho$), we must continue traversing the predicate space and retrain models. Motivated by the above discussions, we propose an incremental DGR discovery algorithm to construct examples that train new models to fit the data only when no existing models can be shared, as shown in Algorithm~\ref{alg: partition}.
\begin{algorithm}[htbp]
\setlength{\intextsep}{1pt}      % 算法在文中时，与上下文本的间距
\setlength{\textfloatsep}{1pt}   % 算法在页面顶部/底部时，与文本的间距
    \caption{DGR Discovery}
    \label{alg: partition}
    \begin{algorithmic}[1]
    \REQUIRE Table $T$ of attribute set $X\subset \mathcal{A}$, target attribute $Y\in \mathcal{A}$, predicate space $P$, error threshold $\rho$
    \ENSURE The DGR-based Examples $E$
    \STATE Example set $E = \emptyset$ 
    \STATE Sharing Model set $M=\emptyset$ 
    \STATE \text{Priority Queue} $Q=\{(R=\emptyset,i(R)=0\}$
    % \STATE Coverage $cov = 0$
    \WHILE {queue $Q$ is not empty}
        \STATE DGR $r= Top(Q)$
        \IF {$\exists m\in M, s. t. error(m(T_r)) \leq \rho_m$}
            \STATE Update $\rho=error(m(T_r))$
            \STATE $E=E\cup{(m,\rho_m,r,T_r)}$
            % \STATE Update coverage $cov\xleftarrow{} cov+\frac{|T_r|}{|T|}$
        \ELSE 
            \STATE $ind(r)=max_{m\in M}\frac{|\{t|t\in T_r\wedge|t.Y-m(t.\mathcal{A})|\leq\rho_m\}|}{|T_r|}$
            \STATE Train $m$ under data $T_r$
            \IF {error $\max_{t\in T_r}error(m(T_r)) \leq \rho_m$}
                \STATE Update $\rho = \max_{t\in T_r}error(m(T_r))$
                \STATE $E=E\cup{(m,\rho_m,r,T_r)}$
                \STATE $M=M\cup{m}$
                % \STATE Update coverage $cov\xleftarrow{} cov+\frac{|T_r|}{|T|}$
            \ELSE 
                \STATE  $P'=TopDown(T_r), P'=\{p_1,...p_n\}\subset P$
                \FOR {$p_i\in P'$}
                    \STATE Update DGR $r_i = r \wedge p_i$
                    \STATE Add to queue $Q = Q\cup {(r_i,ind(r))}$
                \ENDFOR
            \ENDIF
        \ENDIF
    \ENDWHILE
    \RETURN $E$
    \end{algorithmic}
\end{algorithm}

%讲述alg1的大致思路，引出下面的两个优化策略
\textbf{Outline of Algorithm~\ref{alg: partition}.}
Our key idea is to partition data into diverse distributions with DGRs. 
For a given model $m$, we have a corresponding set of DGRs derived from all data it has successfully shared. The DGR with the highest $ind(r)$ value, being the most representative, is used to initialize the prompt in Figure~\ref{fig:DSDG workflow}, while the rest are incorporated as in-context learning examples.
% For the current model $m$, 
% the DGR with the highest $ind(r)$ value is the most representative, which will be utilized to initialize the prompt in Figure~\ref{fig:DSDG workflow}, while the remaining DGRs are incorporated into the in-context learning examples.\lz{what do you mean? clarify it}
We introduce the incremental discovery of DGRs, including
\textsf{model sharing} for Line 7-10, and \textsf{top-down
DGR construction} for Line 12-23. The details are as follows.
% Next, we will introduce these strategies in detail.

\subsubsection{\textbf{Model Sharing Strategy}}\label{sec: Model Sharing}
To reduce model retraining times, for each new DGR $r$, we share a trained model $m$ to build a new example $(m,\rho_m,r,T_r)$ if {error $\max_{t\in T_r}error(m(T_r)) \leq \rho_m$}. By reusing models across DGRs with identical prefixes, we avoid repeated training on similar data, thus speeding up the search process. The effectiveness of this strategy is shown in Example~\ref{example: model sharing}.

% As shown in Example~\ref{example: model sharing}, by sharing models corresponding to DGRs with identical prefixes across different examples, we can speed up the search by avoiding learning the same model multiple times over different parts of the data. 

\begin{example}\label{example: model sharing}
    Take the physical examination~\cite{DBLP:journals/jamia/YadavKCPF17} as an example, we could get DGR $r_1: (height>185cm)\wedge(weight>70kg)$ for a male with the example $e_1:(m,\rho_m, r_1, T_{r_1})$. As we refine DGRs, we find another DGR $r_2: (height>185cm)\wedge(VC>3500ml)$. Both of the DGRs might describe the same group of people, but they could be partitioned into different examples due to the order of searched predicates. If the partitioned subset $T_{r_2}$ satisfies $\max_{t\in T_{r_2}}error(m(T_{r_2})) \leq \rho_m$, then DGR $r_2$ can share the same model $m$ in the example $e_1$.
\end{example}

Further, we have the following conclusion for the correctness of \textsf{model sharing}.

\begin{proposition}[Model Sharing]\label{prop: model sharing}
    If the model $m_1$ in DGR-based example $e_1:(m_1,\rho_1,r_1, T_{r_1})$ could be shared by instance $T_{r_2}$ by applying a predicate $A \phi \delta$, then there exists an example $e_2$ : $(m_2, \rho _2, r_2,T_{r_2})$ s.t.  $e_3:(m_3,\rho_3,r_3, T_{r_3})$ with $m_3= m_1, r_3= (r_2\wedge(A\phi\delta))$ could be reached.
\end{proposition}
% \begin{proof}
%     Consider any tuple $t$ with $t\models e_1$ and $t\models e_2$. If $t\models r_3 = r_1\vee r_2$, there exists some conjunction $C\in r_3$ having $t\models C$. If $C\in r_1$ then $t \models r_1$. It follows $error(m,T_{r_1})<\rho$ since $t\models e_1$. Thus, we have $t\models e_3$. Otherwise, when $C\in r_2$, following a similar proof, we also have $t\models e_3$. To conclude, $e_1, e_2$ imply $e_3$.
% \end{proof}

Proposition~\ref{prop: model sharing} actually points out the existence of a model for a data partition when the model sharing conditions are satisfied. In our Example~\ref{example: model sharing}, if the model can be shared, we only need to change the DGR and the corresponding partitioned data in the example without extra model retraining.

\textbf{Concise DGR-based Examples.} Based on Proposition~\ref{prop: model sharing}, we can aggregate examples with the same initial-stage reasoning steps to build a more concise set of examples. Therefore, we define, state, and prove the soundness of the inference rules for DGR-based examples as follows.

\begin{proposition}[Fusion]\label{prop: fusion}
    Given DGR-based examples $e_1:(m,\rho, r_1, T_{r_1}), e_2:(m,\rho, r_2, T_{r_2})$, $e_1$ and $e_2$ imply a more concise $e_3:(m,\rho,r_3, T_{r_3})$, where $r_3=r_1\vee r_2$.
\end{proposition}

Proposition~\ref{prop: fusion} means that for any tuple $t$ satisfies $e_1$ and $e_2$, if the DGR $r_3=r_1\vee r_2$, then $t$ satisfies $e_3$. In this way, $e_1$ and $e_2$ can be combined into a new DGR-based example $e_3$.

\begin{proposition}[Generalization]\label{prop: generalization}
    If $\rho_2 \geq \rho_1$, then $e_1:(m,\rho_1,r, T_{r})$ implies $e_2:(m,\rho_2,r, T_{r})$.
\end{proposition}
% \begin{proof}
% % [Proof of Proposition~\ref{prop: generalization}]
%     Consider any tuple $t$ with $t\models e_1$. If $t\models r$, Line 14 in Algorithm~\ref{alg: partition} must be satisfied. Thus, we have {$error(m(t)) \leq \rho_1 \leq \rho_2$}. It follows $t\models e_2$. Otherwise, it is natural to reach $t \models e_2$ when $t\not \models r$. To conclude, $e_1$ implies $e_2$.
% \end{proof}

Proposition~\ref{prop: generalization} further considers a tolerance on the error threshold, especially when applying Proposition~\ref{prop: fusion} on two DGR-based examples with different threshold $\rho$. As a result, Proposition~\ref{prop: fusion} and Proposition~\ref{prop: generalization} motivate DATE to build a concise set of DGR-based examples. 

\subsubsection{\textbf{Top-Down DGR Construction}}\label{sec: Top-Down DGR Construction}
In this part, we consider how to construct the predicate space for DGR discovery. Based on Proposition~\ref{prop: model sharing}, we should give priority to building DGRs with identical prefixes to enable better model sharing. Motivated by tree-based model learning~\cite{crr}, top-down traversing may find a model $m$
on each part of the data specified by the conjunction, and refine DGR by adding predicates. In this way, we can maximally retain DGRs with identical prefixes. 

\textbf{Selection of Predicates.} For predicates in the same depth, we prioritize selecting nodes with lower Gini indices~\cite{breiman1984classification} to obtain data with higher purity. However, it is unrealistic to visit all nodes in the decision tree.
% Use all the nodes in the decision tree to build DGRs. %As the depth increases, the importance of nodes decreases successively.
Under the limited model training budget, Theorem~\ref{prop: capacity of predicates} ensures the probability that the model will share at least one constructed DGR as follows.

\begin{theorem}\label{prop: capacity of predicates}
To ensure the probability of at least one constructed DGR being shared, the selected predicates space $P'$ in Algorithm~\ref{alg: partition} should satisfy:
    \begin{equation*}
            |P'|\geq \lceil(1-ind(r))|T_r|\rceil\
    \end{equation*}
\end{theorem}
% \begin{proof}
% % [Proof of Theorem~\ref{prop: capacity of predicates}]
%     Given $ind(r)$ as the sharing probability of DGR $r$, we adopt a worst-case assumption: each predicate covers at most one point in $T_r$. Then, at most $|T_r| \cdot ind(r)$ points can produce shared DGRs, while $(1-ind(r))|T_r|$ cannot. Therefore, $|P'|$ must exceed this latter count, i.e., $|P'| > (1-ind(r))|T_r|$.
% \end{proof}

Based on Theorem~\ref{prop: capacity of predicates}, considering $r$ with index $ind(r)$ in a decreasing order could maximize the probability of \textsf{model sharing} and discovery of DGRs by fewer conjunction splits.

% \subsubsection{\textbf{Early Stop Strategy}}\label{sec: Early Stop}\lz{delete}
% Among Line 13-24, when no model can be shared in $M$, we try to find a new model $m$ to satisfy the error threshold $\rho$. However, if we cannot find a suitable model for the smallest data part, the worst case is indeed a model per tuple, which is already overfitting. Therefore, in Line 27-29, we control the coverage of partitioned data $cov =\sum_{r\in R}\frac{|T_r|}{|T|}$, and end the algorithm when $cov \geq 80\%$ or the partitioned data is less than 1\%. 

\textbf{Time Complexity. }
From Theorem~\ref{prop: capacity of predicates}, the worst case in Algorithm 1 is to train a new model in Line 13 for each tuple. As the top-down searching strategy builds predicates, DGR discovery needs $O(|T|log|T|)$ to construct the nodes. Thus, the time complexity of Algorithm 1 is $O(|T|^2log|T|)$.

\section{Distribution-Specific Generation.}\label{sec: Distribution-Specific Generation}
In this section, we focus on the generation guided by DGR-based LLM reasoning. To address the overfitting problem in Challenge 2, we first capture specific dependencies for the current distribution and generate diverse data through the DGR-guided iterative data generator in Section~\ref{sec: DGR-Guided Iterative Data Generator}. After that, to select the diversity-quality balanced data from numerous generated candidates, we propose a MAB-based sampling strategy with an error bound in Section~\ref{sec: MAB}.
\vspace{-1em}

\subsection{DGR-Guided Iterative Data Generator. }\label{sec: DGR-Guided Iterative Data Generator}
In this section, we formally introduce the generation process of our DATE framework. We propose two key strategies to make LLM reasoning focus on a specific distribution, and formalize them into Algorithm~\ref{alg: generation}. The details are as follows.

\textbf{DGR-guided Generation with Decision Tree Reasoning. }
A natural approach to handling heterogeneous tabular data is to partition the generation process by distribution, employing separate model calls for each. Intuitively, we therefore split the set of DGR-examples according to their distinct underlying distributions, each represented by a unique predictive model. For each resulting partition, we then extrapolate from its DGRs to iteratively generate diverse data, guided and refined by the feedback from decision tree improvements. This process enables the LLM to learn and apply distribution-specific knowledge, continually enhanced by optimizing the DGRs that represent the current data distribution. The detailed procedure is outlined in Algorithm~\ref{alg: generation}.
\begin{algorithm}[htbp]
    \caption{DGR-Guided Data Generation}
    \label{alg: generation}
    \begin{algorithmic}[1]
    \REQUIRE DGR-based examples $E=\{e_i \mid e_i=(m,\rho_m,r,T_r)\}$, shared model pool $M=\{m_1,\dots,m_n\}$, max iterations $I$
    \ENSURE Generated data $T^G=\{(m, T^G_m)\mid m\in M\}$
    \STATE Initialize $T^G_m = \emptyset$ for all $m\in M$
    \STATE Partition $E$: $E_m = \{e\in E \mid e.m = m\}$ for all $m\in M$
    \FOR{each $m \in M$}
        \FOR{$i=1$ to $I$}
            \STATE Partitioned Data $T_m=\bigcup_{r\in E_m} T_{r}$
            \STATE Generated data pool $G_m = \textsc{LLM\_Generate}(E_m)$
            \STATE $\Pi_m = \{\pi(\hat{t}) \mid \pi(\hat{t})=\textsc{Path}(m,\hat{t}),\; \hat{t}\in G_m\}$
            % \STATE // Group candidates by path
            \STATE $\mathcal{H}_m = \{H_k \mid H_k=\{\hat{t}\in G_m:\pi(\hat{t})=k\}\}$
            \FOR{each group $H_k \in \mathcal{H}_m$}
                \IF{$\forall \hat{t}\in H_k:\ error(m,\hat{t})\leq \rho_m$}
                    % \STATE // Accuracy Gain Brought by $H_k$ 
                    \STATE Derive $k^{th}$ DGR $r_k$ from path $\Pi_m$.
                    \STATE $\Delta_k=error(m,T_m)-error(m,T_m\cup H_k)$
                    % \STATE // Build new example for path-specific DGR
                    \STATE $e_k' = (m,\rho_m-\Delta_k,r_k, H_k)$ 
                    \STATE Update $E_m = E_m \cup \{e_k'\}$
                    \IF{$\Delta_k > 0$}
                        \STATE Update $T^G_m = T^G_m \cup H_k$
                    \ENDIF
                \ENDIF
            \ENDFOR
            \STATE $\{r^{new}\} = \textsc{LLM\_RefineRules}(E_m, \{e_k'\})$ 
            \FOR{each $r^{new}$}
                \STATE Update $G_m = G_m\cup \textsc{LLM\_Generate}(r^{new})$
            \ENDFOR
        \ENDFOR
    \ENDFOR
    \RETURN $T^G$
    \end{algorithmic}
\end{algorithm}

\textbf{Outline of Algorithm~\ref{alg: generation}.}
We devise two primary strategies to address the prior bias problem: Decision Tree Path Reasoning to highlight historical predicates in Line 7-9 and DGR Generator Optimization to refine existing DGR in Line 19. Next, we will introduce the two strategies in detail.
% We briefly outline Algorithm~\ref{alg: generation} as follows. To facilitate distribution-specific generation, Line 2 divides the examples according to the sharing model. For the current subset of examples $E_m$, the Decision Tree path reasoning strategy partitions the generated data, which are then scored and organized into new examples (Line 6-16). Based on a comparative analysis of the new and old examples by DGR Generator Optimization, we refine and generate new DGRs and associated data (Line 22-24). We iterate this process in $I$ rounds to simulate the LLM's reflective mechanism. Next, we will introduce the two strategies in detail.

\subsubsection{\textbf{Decision Tree Path Reasoning Strategy}}\label{sec: Decision Tree Path Reasoning Strategy}
% 决策树路径推理
To split long-context from numerous examples and enable distribution-specific generation, Algorithm~\ref{alg: generation} begins by dividing examples into $\{E_m\}$ according to the sharing model $m$, and generates a data pool $G_m$ as candidates. For each record in $G_m$, the Decision Tree model $m$ analyzes its prediction path, and partitions all records with the same path $\pi(\hat{t})=k$ into a new subset $H_k$ (Line 6-9). Intuitively, the decision tree reasoning extracted by $m$ provides valuable insights learned from the entire training dataset. It explicitly \textbf{highlights the historical predicates} for better LLM reasoning in the early stage. We then evaluate the quality of generated records in Line 11, where the error threshold $\rho_m$ must be satisfied. 
% Additionally, we use the validation set of partitioned data $T_m=\bigcup_{r\in E_m} T_{r}$ to evaluate generated data with score  $\Delta_k$, ensuring the LLM learns to generate data that improves

Additionally, we evaluate score $\Delta_k$ of generated data on the partitioned validation set $T_m=\bigcup_{r\in E_m} T_{r}$ to ensure the LLM learns to generate data that improves
% as follows.
% \begin{equation*}
%     \Delta_k=error(m,T_m)-error(m,T_m\cup H_k)
% \end{equation*}
% We monitor $\Delta_k$ to ensure the LLM learns to generate data that improves 
validation performance, rather than merely satisfying the threshold $\rho_m$. As the feedback for LLM reasoning, the path $\pi_k$ with data $H_k$ can be seen as a new DGR with the partitioned data. Thus, we can derive $k$ a new DGR-based example set $e_k' = (m,\rho_m-\Delta_k,\pi_k, H_k)$ in Line 14-18 for more diverse data generation.

\subsubsection{\textbf{DGR Generator Optimization Strategy}}\label{sec: DGR Generator Optimization Strategy}
% 大模型分析DGR中差异
Based on the feedback from \textsf{decision tree reasoning}, Algorithm~\ref{alg: generation} then seeks to optimize existing DGRs for guiding LLM to generate data that can improve the performance in the validation set. Our optimization problem is formalized as follows:

\begin{problem}[DGR Optimization]\label{problem: optimization}
Given the shared model $m$, validation set $T_m$, the DGR generator optimization strategy aims to find a new DGR $r$ satisfying:
    \begin{gather}\nonumber
        \min_r error({m^*},(T_{m_{val}}\oplus r)\\
        \mathrm{subject~to}\quad m^*=\arg\min_m error(m,(T_{m_{train}}\oplus r)) \nonumber
    \end{gather}
where $T_m\oplus r$ denotes the data augmented with the new records generated by $r$. In summary, we optimize the rule $r$ to achieve the best validation score measured on $T_{m_{val}}\oplus r$ with the model $m^*$ trained to
minimize the loss on $T_{m_{train}}\oplus r$.
\end{problem}

However, such bi-level optimization is often computationally demanding, as it involves computing gradients through the optimization of $m^*$. Recently, optimization using LLMs has proven to be an effective tool~\cite{yang2024large}, outperforming over traditional solvers in tasks like prompt optimization, where its capacity to iteratively refine outputs based on historical trajectory allows it to discover high-performing strategies that are non-obvious to hand-crafted heuristics. Motivated by this, we leverage LLMs to reason the relationship between $E_m$ and ${e'_k}$, prompting it to propose a new DGR $r^{new}$ that is absent from $E$ and would improve the scores achieved in previous iterations (Line 22). The detailed prompt is available in our code~\cite{github}.

%写几句话引出sampling
\textbf{Generating Multiple Candidate Data.}
DGR Generator Optimization proceeds for $I$ iterations to generate multiple useful DGRs and data. For example, after we find that the predicate $(height > 185cm)$ contributes to better performance, an additional DGR can be generated by conjoining it. Algorithm~\ref{alg: generation} iteratively generates new DGRs and corresponding data, until the validation score no longer improves or 
reaches $I$ rounds. All optimized DGRs and diverse data $T_m^G$ are preserved with the validation score. Next, we will introduce how we select the diversity-quality balanced data from numerous candidates.

\textbf{LLM-Call complexity.}
From Theorem~\ref{prop: capacity of predicates}, the worst case is $M = |T|$. As each refinement step produces exactly one new $r_{new}$ and triggers a single LLM invocation, 
% the algorithm makes $M\times I$ LLM calls across M models and I iterations. Thus, 
the LLM-call complexity of Algorithm 3 is $O(I|T|)$, which simplifies to $O(|T|)$ when I is treated as a constant.

\subsection{MAB-based Generated Data Sampling}\label{sec: MAB}
As the final step of generation in DATE, we integrate the iteratively generated data for various distributions to obtain the final result. In this section, we first clarify that greedy-based selection is unreliable for heterogeneous data, and then propose a Multi-Arm-Based (MAB) sampling algorithm to select diversity-quality balanced data for each distribution represented by model $m$.

\textbf{Greedy-based Selection is not Suitable for Heterogeneous Data.}
To integrate numerous generated data $T^G$ from Algorithm~\ref{alg: CSAR} into the final augmented result, existing LLM-based generators such as EPIC~\cite{EPIC} and GReaT~\cite{GReaT} typically rely on greedy selection of the “best” iteration according to the validation score. \uline{However, little theoretical work has been done to reveal the heterogeneous nature of tabular data, which poses a key problem as the synthetic data selection defined in}~\cite{CLLM}. We observe that the data selected from the best-performing validation round cannot be guaranteed to be optimal over the entire validation set. We formalize this key observation as Theorem~\ref{greedy}.

\begin{theorem}\label{greedy}
    % For heterogeneous tabular data composed of different distributions as defined in~\cite{JMLR:v20:13-580,8475006},
    % % , which can be viewed as coming from multiple sources with feature spaces that have the same dimension but different distributions, 
    % the synthetic data selection problem does not possess the greedy-choice property.

    Synthetic data selection for heterogeneous data defined in~\cite{JMLR:v20:13-580,8475006}, whose feature space is composed of different distributions, lacks the greedy-choice property.
\end{theorem}

% \begin{proof}
% % [Proof of Theorem~\ref{greedy}]
% Suppose that the greedy algorithm first selects a subset \( H_i \) that maximizes \( \Delta_i \). However, the local gain \( \Delta_k \) for each specific distribution \( \mathcal{D}_k \) is evaluated on its local partition \( T_m \), while the overall model performance is non-additive across heterogeneous distributions. Therefore, even if \( H_i \) is a component of the global optimum, selecting the best combination from the other distributions given \( H_i \), does not necessarily preserve optimality. To conclude, the greedy-choice property does not hold for the generated data selection.
% \end{proof}

Based on Theorem~\ref{greedy}, we find that choosing only the best iteration would discard the diversity across iterations. In fact, data diversity possesses the potential to enhance overall data performance~\cite{diversity}, while we still need to select high-quality data to ensure the effectiveness. The technical challenge is to balance this exploration and exploitation. Inspired by this, we propose a MAB-based solution as follows.

\textbf{Novelty of Sampling for Iterative Generation.} We make contributions to proposing simple yet effective solutions never explored in iterative generation studies. We present a novel view of the diversity for generated data based on DGRs, and adapting Multiple Arm Identification (MAI) strategies to select data from iteratively generated candidates. Our sampling algorithm is particularly flexible and could be applied to numerous real-world problems that require a diversity-quality tradeoff, such as core-set selection~\cite{Quad,Perplexity} and other iterative data generation methods~\cite{2014PrivBayes,Tabby,Synthesizers,GReaT,EPIC}.

\textbf{Multi-armed Bandit (MAB).} MAB is a slot machine with $K$ arms$\{1,..,K\}$. An agent decides which arm to pull in each of $n$ total rounds. When pulled, arm $i$ returns a random reward from an unknown reward distribution $D_i$ specific to arm $i$.

\textbf{Multiple Arm Identification (MAI).} %Unlike the traditional goal where the agent maximizes the total reward,
%or minimize the regret
In \textsl{Multiple Arm Identification (MAI)}, the agent identifies a subset of the arms $\{J_1,...\} \subset \{1,..,K\}$ corresponding to a specified criteria~\cite{b2}.

\textbf{Model as MAI.} 
Choosing a new DGR $r_k$ with LLM-generated data $H_K$ for the whole validation set can be viewed as Multiple Arm Identification, where each arm is a DGR $r_k$, and the reward distribution is made up from the diversity using this DGR and the validation score of $H_k$.

\textbf{The MAB-based Sampling Solution.}
Deriving the best algorithm for a specific case of MAI is an open problem~\cite{MAB}. We adopt the SAR algorithm for our problem as shown in Algorithm~\ref{alg: CSAR}. Here, each arm is a DGR $r_k$, which represents the corresponding generated data $H_k$. Selecting the $k$-th arm (Line 9) is sampling a subset of DGR containing the $k$-th DGR and recording the performance of the shared model $m$ using subset $\sum{H_k}$ into the expectation approximation. Accepting the $k$-th arm $e_k$ (Line 7) means deciding its expectation is higher than the best validation score.

\begin{algorithm}[htbp]
	\caption{MAB-based Data Sampling (MDS)}
    \label{alg: CSAR}
	\begin{algorithmic}[1]
		\REQUIRE The shared model $m$ with partitioned data $T_m$, original examples $E_m$, newly generated data for model $m$: $T_m^G=\{e_i\}_{i=1}^K$, where $e_k=(m,\rho_k,r_k,H_k)$, budget $n$, weight $\alpha\in(0,1)$
        \ENSURE The Accepted arms $A_K$
        \STATE $n_0 = 0$; $A_K = \emptyset$; $A_1 = \{1,\dots,K\}$
        \STATE The best score $bs=0$
        \FOR {arm $e_k \in T_m^G$}
            \STATE $div_k = div(r_k | E_m)$
            \STATE $u_k = \alpha\cdot (1-\rho_k) + (1-\alpha)\cdot \mathrm{div}_k$
        \ENDFOR
		\FOR {$k=1,2,\dots,K-1$}
            \STATE $n_k=\left\lceil \frac{1}{\overline{\log}K}\cdot \frac{n-K}{K+1-k} \right\rceil$
            
            \STATE Select the arm $i\in A_k$ for $n_k-n_{k-1}$ rounds 
		    \STATE $e_{i_k} = \arg\max_{i\in A_k} |u_{i_k}|$
            \STATE $A_{k+1} = A_k \setminus \{i_k\}$
            \STATE Update $u_{i_{k}}$ by $m(H_{i_k}, T_m)$
            \IF {$u_{i_k}\geq bs$}
                \STATE $bs = u_{i_k}$
		        \STATE Accept arm $A_K = A_K \cup \{i_k\}$ 
            \ENDIF
        \ENDFOR
        \RETURN $A_K$
	\end{algorithmic}
 \label{alg:CSAR-PathSim}
\end{algorithm}

\textbf{The Design of Diversity and Quality in MDS.}
One of the key challenges of Algorithm~\ref{alg: CSAR} is to evaluate the diversity and quality of the arm $e_k$. On the one hand, we retain $\rho_k$ from Line 15 in Algorithm~\ref{alg: generation} as the quality. On the other hand, we formalize the diversity of the arm $e_k$ as follows.
\begin{definition}[Data Diversity for DGR-based Examples]
    Given the MAB arm $e_k=(m,\rho_k,r_k,H_k)$, the original example set $E_m=\{e_j | e_j = (m, \rho_j, r_j, T_j)\}_{j=1}^J$ for shared model $m$, the diversity for DGR-based examples $e_k$ for the current shared model $m$ is the weighted sum of similarity between $r_k$ and $E_m$ as follows:
    \begin{equation*}
        div(e_k | E_m) = \sum_{e_j \in E_m} w_j \cdot Overlap(r_k, r_j)
    \end{equation*}
    where $w_j = |T_j| / \sum_{l=1}^J |T_{l}|$ is the weight for each example $e_j$, and $Overlap(r_k, r_j) = \frac{|P(r_k) \cap P(r_j)|}{|P(r_k) \cup P(r_j)|}$ represents the interval overlap compatibility~\cite{overlap} between DGR $r_k$ and $r_j$.
\end{definition}

% To select the most optimal data across the entire validation set, MDS considers the diversity of $r_k$ by comparing similarities with the entire DGRs in the original example set $E_m$, which actually constitutes a mixture distribution of the original validation set. We introduce $\alpha$ to establish a normalized optimization objective in Line 5. In our experiments, we set $\alpha=0.8$ for all datasets, corresponding to the best experimental results obtained.

To select the most optimal data across the entire validation set, MDS accounts for the diversity of \( r_k \) by measuring its similarity with all DGRs in the original example set \( E_m \), which essentially forms a mixture distribution of the original validation set. We introduce a parameter \( \alpha \) to formulate a normalized optimization objective in Line 5. In our experiments, \( \alpha = 0.8 \) was set for all datasets as the best experimental results.

\textbf{Iterative Optimization.}
MDS iteratively selects the new arm $e_{i_k}$ in Line 9-11.
The potential rewards $u$ of unselected arms will increase according to the Upper Confidence Bound (UCB) algorithm~\cite{MAB}. Also, the diversity for the subsequent arm $e_k$ will incorporate accepted arms $A_K$ as part of the existing distribution to calculate similarity as follows.
\begin{equation*}
    div_k=div(r_k|E_m\cup A_K)
\end{equation*}
This process continues until the global best score $bs$ no longer improves or reaches the budget $n$.

\textbf{Error and Time Cost.} 
Given an acceptable error bound $e_n\leq \delta$, Proposition~\ref{prop: error bound} shows that MDS needs to train $2\overline{log}K*S*ln(\frac{2K^2}{\delta})+K = O(ln(\frac{1}{\delta}))$ models online. 

\begin{proposition}\label{prop: error bound}
    The probability of error in MDS satisfies:
        \begin{equation*}
                e_n \leq 2K^2exp(-\frac{n-K}{2\overline{log}K*S}) \nonumber
    \end{equation*}
    \vspace{-1em}
\end{proposition}
% The probability of error satisfies:
where $S = max_{i\in \{1,..,K\}} i*(|\mu_i|)^{-2}$, $\overline{log}K = \frac{1}{2} + \sum^K_{i=2}\frac{1}{i}$.
If we limit the quantity of generated data for each round in Algorithm~\ref{alg: generation}, $k$ would be a constant. In this case, Algorithm~\ref{alg: CSAR} requires a $n=O(ln(\frac{1}{\delta}))$ time cost on sampling.

% \begin{theorem}[Optimality of MDS for Subset Selection]
% MDS achieves the optimal error rate for subset selection.
% \begin{proof}
% For subset selection with top-$m$ arms, the best lower bound achieved by Successive Elimination is: 
% \[
% e_n^{\text{lower}} \geq \exp\left(-\frac{n}{H_{\text{subset}}} + o(n)\right)
% \]
% where 

% $H_{\text{subset}} = \max\left\{\max_{i \leq m} \frac{1}{(\mu_i - \mu_{m+1})^2}, \max_{j \geq m+1} \frac{1}{(\mu_m - \mu_j)^2}\right\}$

% Since $S = \Theta(H_{\text{subset}})$ and $2\overline{\log K} = \Theta(\log K)$, we have:
% \[
% \frac{1}{2\overline{\log K} \cdot S} = \Theta\left(\frac{1}{H_{\text{subset}} \cdot \log K}\right)
% \]

% To conclude, MDS matches the optimal rate achieved by SOTA algorithms like Successive Elimination.
% \end{proof}
% \end{theorem}

\section{Experiments}
\label{sec:experiemnt}
We conduct experiments on real-world Data Generation benchmarks to evaluate DATE performance. In this section, we first introduce experimental settings in Section~\ref{sec:section5.1}, followed by an end-to-end performance evaluation in Section~\ref{sec: end-to-end}, a detailed ablation study in Section~\ref{sec: ablation}, and a parameter analysis in Section~\ref{sec: Sensitivity Analysis}. Empirically, DATE outperforms the state-of-the-art LLM generator and can provide reliable data for better Direct Preference Optimization (DPO) performance.

\subsection{Experimental Settings}\label{sec:section5.1}

\textbf{Datasets.} 
When selecting datasets, it is important to consider the heterogeneous nature of tabular data~\cite{STUNT} and ensure coverage of both categorical and numerical features, as well as both classification and regression tasks. We use the popular datasets in OCTree~\cite{fernando2024promptbreeder}, including 8 classification datasets and 2 regression datasets. We split each dataset and use 60\% for training, 20\% for validation, and 20\% for testing. 

\textbf{Parameter setting.}
We set the default parameters as follows: error threshold $\rho=0.05$ in Algorithm~\ref{alg: partition}, the iteration round $I=3$ in Algorithm~\ref{alg: generation}, and quality weight $\alpha=0.8$ in Algorithm~\ref{alg: CSAR}. The value of the budget for MAB-based sampling is initialized as 200, according to available computing resources. The sensitivity analysis can be found in Section~\ref{sec: Sensitivity Analysis}

\textbf{Baselines.} 
To validate the effectiveness of DATE, we select one representative GAN-based tabular data generation solution and three SOTA LLM-based solutions for comparison. Furthermore, we choose four data selection methods to compare the performance of MDS.

$\bullet$ CTGAN~\cite{CTGAN} is a GAN-based model designed for tabular data generation, which leverages conditional GANs to effectively capture the underlying distribution of complex datasets while preserving the relationships between features.

$\bullet$ GReaT~\cite{GReaT} is a finetuning-based solution, which incorporates domain knowledge to enhance the quality of samples generated by GPT 2.5.

$\bullet$ CLLM~\cite{CLLM} is a prompt-based generator which leverages LLMs' prior knowledge and learning dynamics-driven curation to address data scarcity in low-data regimes.

$\bullet$ EPIC~\cite{EPIC} is a prompt-based SOTA tabular data generator, which leverages balanced, grouped data samples and variable mapping to guide LLMs in addressing class imbalance.

$\bullet$ DATE (w/ examples) is a variant of DATE that employs the DGR-based prompt in Algorithm 1 for in-context learning. It directly instructs the LLM to generate new data, without involving any optimization strategies from Algorithm~\ref{alg: generation} or~\ref{alg: CSAR}.

$\bullet$ Quad~\cite{Quad} is a scalable data selection framework that integrates attention-layer influence computation with MAB-driven sampling for diversity-aware pretraining.

$\bullet$ Perplexity Correlations~\cite{Perplexity} is a SOTA pretraining data selection solution that
% leveraging public LLMs' causal language modeling loss and downstream performance to 
estimates perplexity-benchmark correlations and selects high-correlation web data for pretraining.

\begin{table*}[ht]
    \centering
    \caption{Effectiveness evaluation of end-to-end solutions on Tabular Data Generation benchmarks. We report \textbf{error} and its \textbf{percentage} increase, along with the \textbf{SYN.} (synthetic data volume), as evaluation metrics. Specifically, the error corresponds to the error rate for \colorbox[HTML]{CBCEFB}{classification tasks} (highlighted in blue) and the MSE for \colorbox[HTML]{FFCCC9}{regression tasks} (highlighted in red).
    The BEST error in each row is shown in \textbf{BOLD}, and the SECOND-BEST error is shown in \underline{UNDERLINE}. }
    \label{table:all performance}
    \setlength\tabcolsep{2pt}
    \setlength{\extrarowheight}{2pt} % 增加行高
    \resizebox{\textwidth}{!}{%
\begin{tabular}{@{}cc|cc|cccc|cccccccc|cccc@{}}
\toprule
\multicolumn{2}{c|}{\textbf{Original Data}}                                                                        & \multicolumn{2}{c|}{\textbf{GAN-based}} & \multicolumn{4}{c|}{\textbf{Fine-tuning based}}                         & \multicolumn{8}{c|}{\textbf{Prompt-based}}                                                                                                                  & \multicolumn{4}{c}{\textbf{Our Method}}                                                   \\ \midrule
                                                                                     &                             & \multicolumn{2}{c|}{CTGAN}              & \multicolumn{2}{c}{GReaT-III}      & \multicolumn{2}{c|}{GReaT-V}       & \multicolumn{2}{c}{EPIC-I}        & \multicolumn{2}{c}{EPIC-II}              & \multicolumn{2}{c}{EPIC-III}             & \multicolumn{2}{c|}{CLLM}         & \multicolumn{2}{c}{DATE-I}            & \multicolumn{2}{c}{DATE}                          \\
\multirow{-2}{*}{Dataset}                                                            & \multirow{-2}{*}{Real Data} & error       & SYN.                      & error     & SYN.                   & error     & SYN.                   & error    & SYN.                   & error            & SYN.                  & error            & SYN.                  & error    & SYN.                   & error         & SYN.                  & error            & SYN.                           \\ \midrule
\cellcolor[HTML]{CBCEFB}                                                             &                             & 28.686      &                           & 26.512    &                        & 26.418    &                        & 23.771   &                        & {\ul 21.690}     &                       & 23.020           &                       & 27.599   &                        & 22.980        &                       & \textbf{20.860}  &                                \\
\multirow{-2}{*}{\cellcolor[HTML]{CBCEFB}Bank-Marketing}                             & \multirow{-2}{*}{28.990}    & 1.05\%      & \multirow{-2}{*}{10000}   & 8.55\%    & \multirow{-2}{*}{1000} & 8.87\%    & \multirow{-2}{*}{1000} & 18\%     & \multirow{-2}{*}{2191} & {\ul 25.18\%}    & \multirow{-2}{*}{791} & 20.59\%          & \multirow{-2}{*}{692} & 4.80\%   & \multirow{-2}{*}{1000} & 20.73\%       & \multirow{-2}{*}{120} & \textbf{28.04\%} & \multirow{-2}{*}{\textbf{341}} \\
\cellcolor[HTML]{CBCEFB}                                                             &                             & 14.295      &                           & 13.567    &                        & 14.542    &                        & 22.547   &                        & 15.000           &                       & \textbf{11.500}  &                       & 13.574   &                        & 13.099        &                       & {\ul 12.775}     &                                \\
\multirow{-2}{*}{\cellcolor[HTML]{CBCEFB}Electricity}                                & \multirow{-2}{*}{13.437}    & -6.39\%     & \multirow{-2}{*}{10000}   & -0.97\%   & \multirow{-2}{*}{1000} & -8.22\%   & \multirow{-2}{*}{1000} & -67.80\% & \multirow{-2}{*}{1195} & -11.63\%         & \multirow{-2}{*}{300} & \textbf{14.42\%} & \multirow{-2}{*}{564} & -1.02\%  & \multirow{-2}{*}{1000} & 2.52\%        & \multirow{-2}{*}{56}  & {\ul 4.93\%}     & \multirow{-2}{*}{\textbf{139}} \\
\cellcolor[HTML]{CBCEFB}                                                             &                             & 21.824      &                           & 21.039    &                        & 21.562    &                        & 20.243   &                        & 20.120           &                       & 19.750           &                       & 37.855   &                        & {\ul 17.150}  &                       & \textbf{16.580}  &                                \\
\multirow{-2}{*}{\cellcolor[HTML]{CBCEFB}MagicTelescope}                             & \multirow{-2}{*}{21.114}    & -3.36\%     & \multirow{-2}{*}{10000}   & 0.35\%    & \multirow{-2}{*}{942}  & -2.12\%   & \multirow{-2}{*}{1000} & 4.12\%   & \multirow{-2}{*}{1300} & 4.71\%           & \multirow{-2}{*}{370} & 6.46\%           & \multirow{-2}{*}{404} & -79.29\% & \multirow{-2}{*}{1000} & {\ul 18.77\%} & \multirow{-2}{*}{54}  & \textbf{21.47\%} & \multirow{-2}{*}{\textbf{155}} \\
\cellcolor[HTML]{CBCEFB}                                                             &                             & 44.350      &                           & 46.189    &                        & 43.116    &                        & 45.467   &                        & \textbf{41.000}  &                       & 44.590           &                       & 45.700   &                        & 44.026        &                       & {\ul 43.098}     &                                \\
\multirow{-2}{*}{\cellcolor[HTML]{CBCEFB}Eye-Movements}                              & \multirow{-2}{*}{45.926}    & 3.43\%      & \multirow{-2}{*}{10000}   & -0.57\%   & \multirow{-2}{*}{176}  & 6.12\%    & \multirow{-2}{*}{1000} & 1\%      & \multirow{-2}{*}{1221} & \textbf{10.73\%} & \multirow{-2}{*}{430} & 2.91\%           & \multirow{-2}{*}{392} & 0.49\%   & \multirow{-2}{*}{1000} & 4.14\%        & \multirow{-2}{*}{33}  & {\ul 6.16\%}     & \multirow{-2}{*}{\textbf{62}}  \\
\cellcolor[HTML]{CBCEFB}{}                    &                             & 36.900      &                           & 35.900    &                        & 35.250    &                        & 35.200   &                        & 41.000           &                       & 35.870           &                       & 36.150   &                        & {\ul 34.720}  &                       & \textbf{34.500}  &                                \\
\multirow{-2}{*}{\cellcolor[HTML]{CBCEFB}{Heloc}} & \multirow{-2}{*}{36.650}    & -0.68\%     & \multirow{-2}{*}{10000}   & 2.05\%    & \multirow{-2}{*}{785}  & 3.82\%    & \multirow{-2}{*}{997}  & 3.96\%   & \multirow{-2}{*}{1178} & -11.87\%         & \multirow{-2}{*}{470} & 2.13\%           & \multirow{-2}{*}{309} & 1.36\%   & \multirow{-2}{*}{1000} & {\ul 5.27\%}  & \multirow{-2}{*}{28}  & \textbf{5.87\%}  & \multirow{-2}{*}{\textbf{62}}  \\
\cellcolor[HTML]{CBCEFB}                                                             &                             & 31.110      &                           & 30.960    &                        & 30.452    &                        & 28.148   &                        & 27.539           &                       & 34.000           &                       & 28.298   &                        & {\ul 25.030}  &                       & \textbf{24.790}  &                                \\
\multirow{-2}{*}{\cellcolor[HTML]{CBCEFB}Credit}                                     & \multirow{-2}{*}{30.900}    & -0.68\%     & \multirow{-2}{*}{10000}   & -0.19\%   & \multirow{-2}{*}{1000} & 1.45\%    & \multirow{-2}{*}{1000} & 8.90\%   & \multirow{-2}{*}{1604} & 10.88\%          & \multirow{-2}{*}{515} & -10.03\%         & \multirow{-2}{*}{564} & 8.42\%   & \multirow{-2}{*}{1000} & {\ul 19.00\%} & \multirow{-2}{*}{28}  & \textbf{19.77\%} & \multirow{-2}{*}{\textbf{149}} \\
\cellcolor[HTML]{CBCEFB}                                                             &                             & 16.404      &                           & 16.622    &                        & 17.325    &                        & 21.008   &                        & 13.860           &                       & 18.560           &                       & 27.332   &                        & {\ul 13.715}  &                       & \textbf{12.915}  &                                \\
\multirow{-2}{*}{\cellcolor[HTML]{CBCEFB}California}                                 & \multirow{-2}{*}{15.605}    & -5.12\%     & \multirow{-2}{*}{10000}   & -6.52\%   & \multirow{-2}{*}{1000} & -11.03\%  & \multirow{-2}{*}{1000} & -34.63\% & \multirow{-2}{*}{1503} & 11.18\%          & \multirow{-2}{*}{502} & -18.94\%         & \multirow{-2}{*}{483} & -75.16\% & \multirow{-2}{*}{1000} & {\ul 12.11\%} & \multirow{-2}{*}{171} & \textbf{17.24\%} & \multirow{-2}{*}{\textbf{301}} \\
\cellcolor[HTML]{CBCEFB}                                                             &                             & 31.122      &                           & 30.914    &                        & 30.540    &                        & 29.968   &                        & 28.380           &                       & \textbf{21.130}  &                       & 42.376   &                        & 27.749        &                       & {\ul 26.836}     &                                \\
\multirow{-2}{*}{\cellcolor[HTML]{CBCEFB}Jannis}                                     & \multirow{-2}{*}{30.158}    & -3.20\%     & \multirow{-2}{*}{10000}   & -2.51\%   & \multirow{-2}{*}{817}  & -1.27\%   & \multirow{-2}{*}{651}  & 0.63\%   & \multirow{-2}{*}{1039} & 5.90\%           & \multirow{-2}{*}{369} & \textbf{29.94\%} & \multirow{-2}{*}{353} & -40.51\% & \multirow{-2}{*}{986}  & 7.99\%        & \multirow{-2}{*}{16}  & {\ul 11.02\%}    & \multirow{-2}{*}{\textbf{41}}  \\ \midrule
CLS-AVG                                                      & 27.847                      & 28.086      & 10000                     & 27.713    & 840                    & 27.401    & 956                    & 28.294   & 1404               & 26.074           & 468               & 26.053           & 470               & 32.361   & 998                    & {\ul 24.809}  & 63                    & \textbf{24.044}  & \textbf{156}                   \\ \midrule
\cellcolor[HTML]{FFCCC9}                                                             &                             & 96.919      &                           & 232.643   &                        & 273.881   &                        & 23.167   &                        & 33.660           &                       & 30.310           &                       & 20.400   &                        & {\ul 7.541}   &                       & \textbf{7.517}   &                                \\
\multirow{-2}{*}{\cellcolor[HTML]{FFCCC9}Bike}                                       & \multirow{-2}{*}{31.242}    & -210.22\%   & \multirow{-2}{*}{2000}    & -644.64\% & \multirow{-2}{*}{5793} & -776.64\% & \multirow{-2}{*}{168}  & 25.85\%  & \multirow{-2}{*}{686}  & -7.74\%          & \multirow{-2}{*}{241} & 2.98\%           & \multirow{-2}{*}{345} & 34.70\%  & \multirow{-2}{*}{1000} & {\ul 75.86\%} & \multirow{-2}{*}{9}   & \textbf{75.94\%} & \multirow{-2}{*}{\textbf{18}}  \\
\cellcolor[HTML]{FFCCC9}                                                             &                             & 59.570      &                           & 59.68     &                        & 34.340    &                        & 20.990   &                        & 30.045           &                       & 30.450           &                       & 21.500   &                        & {\ul 20.785}  &                       & \textbf{11.789}  &                                \\
\multirow{-2}{*}{\cellcolor[HTML]{FFCCC9}Boston}                                     & \multirow{-2}{*}{22.272}    & -167.47\%   & \multirow{-2}{*}{2000}    & -167.96\% & \multirow{-2}{*}{3475} & -54.18\%  & \multirow{-2}{*}{101}  & 5.76\%   & \multirow{-2}{*}{482}  & -34.90\%         & \multirow{-2}{*}{161} & -36.72\%         & \multirow{-2}{*}{241} & 3.47\%   & \multirow{-2}{*}{1000} & {\ul 6.68\%}  & \multirow{-2}{*}{10}  & \textbf{47.07\%} & \multirow{-2}{*}{\textbf{16}}  \\ \midrule
REG-AVG                                                      & 26.757                      & 78.245      & 2000                      & 146.162   & 4634                   & 154.111   & 135                    & 22.078   & 584                    & 31.853           & 201                   & 30.38            & 293                   & 20.95    & 1000                   & {\ul 14.163}  & {\ul 9.5}             & \textbf{9.63}    & \textbf{17}                    \\ \bottomrule
\end{tabular}%
}
\end{table*}

\textbf{Evaluation Metrics and Environment.} 
We use deepseek-r1~\cite{deepseek} for generation in Algorithm~\ref{alg: generation}. Each baseline is evaluated using either DecisionTreeRegressor or DecisionTreeClassifier in scikit-learn~\cite{pedregosa2011scikit}, depending on its tasks. We report the test error rates (\%) for classification tasks, Mean Squared Error (MSE) for regression tasks, and the scale of generated data (SYN.) for cost on all datasets to ensure a fair comparison~\cite{STUNT, GReaT}. 

\subsection{End-to-end performance}\label{sec: end-to-end}
We conduct end-to-end comparisons on all datasets for classification tasks and regression tasks separately. All experiments were based on the open-source code of the baselines and adopted the parameters recommended in the original paper. Our results are shown in Table~\ref{table:all performance}

\textbf{Comparison with GAN-based Solutions.}
We first compare DATE with the state-of-the-art GAN-based solution. As shown in Table~\ref{table:all performance}, DATE significantly surpasses state-of-the-art GAN-based methods such as CTGAN across all datasets, a trend also observed in other LLM-based approaches like EPIC~\cite{EPIC} and CLLM~\cite{CLLM}. This performance gap stems from fundamental limitations of GANs: the adversarial zero-sum game between generator and discriminator often causes mode collapse, where only a limited set of sample types are produced, sacrificing diversity and failing to capture heterogeneous data distributions. In contrast, DATE addresses this by partitioning the original data into clearer, high-quality distributions to facilitate targeted generation, while leveraging LLMs to progressively expand and diversify examples through stable denoising-based objectives. This enables DATE to comprehensively cover distribution modes, inherently avoiding mode collapse and achieving superior accuracy with enhanced diversity.

\textbf{Comparison with Finetuning-based Solutions.}
We compare DATE with the representative finetuning-based solution GReaT~\cite{GReaT}. We apply the same experiment configuration as the original paper and separately present the results of GReaT after three rounds (GReaT-III, which has the same rounds as ours) and after five rounds (GReaT-V) in Table~\ref{table:all performance}. DATE significantly outperforms both of them among all datasets. It is worth noticing that on certain datasets such as California, Electricity, and MagicTelescope, GReaT-V performs worse than GReaT-III, which suggests that excessive rounds may lead to overfitting caused by data hungry. In Section~\ref{sec: Sensitivity Analysis}, we also conduct a parameter study on the iteration round $I$ to evaluate the robustness of DATE. In the Appendix, we provide a detailed discussion on why DATE can outperform fine-tuning based solutions in heterogeneous settings.

\textbf{Comparison with Prompt-based Solutions.}
We apply the same iteration rounds $I=3$ for iterative prompt-based solutions EPIC and CLLM. DATE outperforms them across most real-world datasets. \uline{Remarkably, with merely \textbf{18} generated data on the ``bike'' regression dataset, DATE achieves a 75\% reduction in MSE. For the Jannis classification dataset (57,580 rows, 55 columns), DATE achieved an 11\% reduction in error rate with only 41 generated samples, reaching the current second-best performance.} From Table~\ref{table:all performance}, we observe that EPIC may also suffer from overfitting. For instance, on the EM, CR and CA, the second iteration consistently outperforms the third, which could be attributed to the catastrophic forgetting or error accumulation during iterative learning. In contrast, DATE remains robust by leveraging \textsf{decision tree reasoning} to prioritize historically effective predicates and filter low-quality data (Algorithm~\ref{alg: generation}, Line 11), thereby promptly correcting the LLM's learning trajectory.

\textbf{Can DATE Understand the Heterogeneous Nature of Real-World Data Better?}
Real-world tabular data has a heterogeneous nature and often lacks semantic descriptions, making it challenging for LLMs to comprehend its semantics. \uline{In Table~\ref{table:all performance}, we observe that some augmented results are even worse than the original performance}, indicating that the baselines struggle to deal with heterogeneous data. Existing baselines tend to rely on analyzing the frequency of each token's occurrence. In this case, data examples in the prompt may become repetitive, making it difficult for the model to distinguish between variables. \uline{It is worth noting that DATE (w/ examples) has achieved the second-best performance on many datasets}, underscoring the importance of exploring data heterogeneity. Instead of learning the occurrence patterns of symbols as in EPIC~\cite{EPIC}, DATE employs the \textsf{decision tree reasoning} to discovery historically effective predicates, and iteratively refines DGRs to narrow down the value ranges through \textsf{DGR optimization}, enabling a layer-by-layer inference of the decision tree structure, allowing DATE to effectively learn contextual semantics and understand the heterogeneous nature of real-world data better.

% DATE learns the patterns of predicates through the \textsf{DGR optimization}. DATE iteratively refines DGRs to narrow down the value ranges, enabling a layer-by-layer inference of the decision tree structure, allowing DATE to effectively learn contextual semantics and understand the heterogeneous nature of real-world data better.

\textbf{Takeaway:}
DATE achieves the best average performance among all SOTA baselines with minimum synthetic data, improving classification and regression tasks by 13.66\% and 64\%, respectively.
% , which validates its effectiveness in identifying mixed distributions and enabling targeted generation. 
The promising performance of DATE (w/ examples) further confirms that exploring data heterogeneity and constructing high-quality examples are crucial for LLMs.

% DATE improves the average performance on classification tasks by 13.66\% and on regression tasks by a substantial 64\%, demonstrating its effectiveness in identifying mixed distributions within real-world data and performing precise, targeted generation. 
% Furthermore, the promising performance of DATE (w/ examples) reveals the importance of exploring the heterogeneous nature of tabular data and constructing high-quality examples for enhancing LLM effectiveness.
\vspace{-1em}
\subsection{Ablation study}\label{sec: ablation}
\begin{figure}[htbp]
    \centering
    \vspace{-1em}
    \includegraphics[width=\columnwidth]{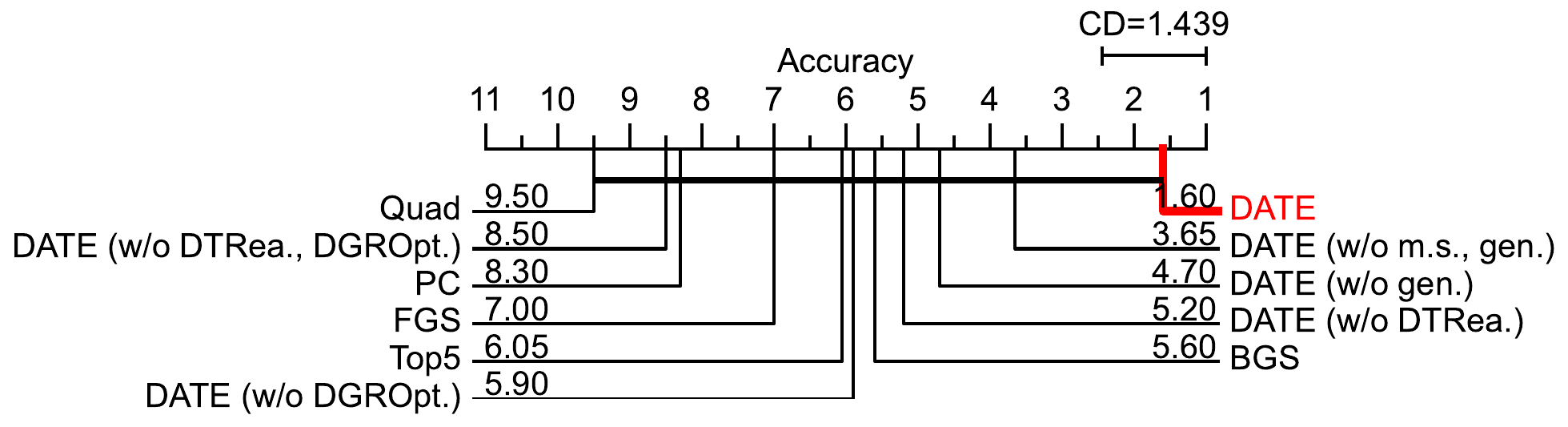}
    \caption{CD diagram for DATE against ablation baselines in terms of accuracy ranking. The significance level $\alpha=0.05$. 
    The top-ranked method is highlighted in \textcolor{red}{\textbf{red}}.
    }
    \label{fig: cd}
\end{figure}

In this section, we depose the modules in the DATE framework and analyze their effectiveness. The accuracy ranking of all compared baselines is shown in Figure~\ref{fig: cd}.

\textbf{Effectiveness of Model Sharing.}
As the foundation of DATE, we first study the \textsf{model sharing} strategy, which accelerates DGR discovery, guides the traversal of predicates to construct specific and concise examples. In Figure~\ref{fig:model_sharing}, we analyze the \textsf{model sharing} strategy for DGR discovery across all datasets. We made the following observations:

\begin{figure}[htbp]
\vspace{-1em}
    \centering
    % 显示整个合并图片
    \includegraphics[width=\columnwidth,clip,
    trim=0 0 0 0,]{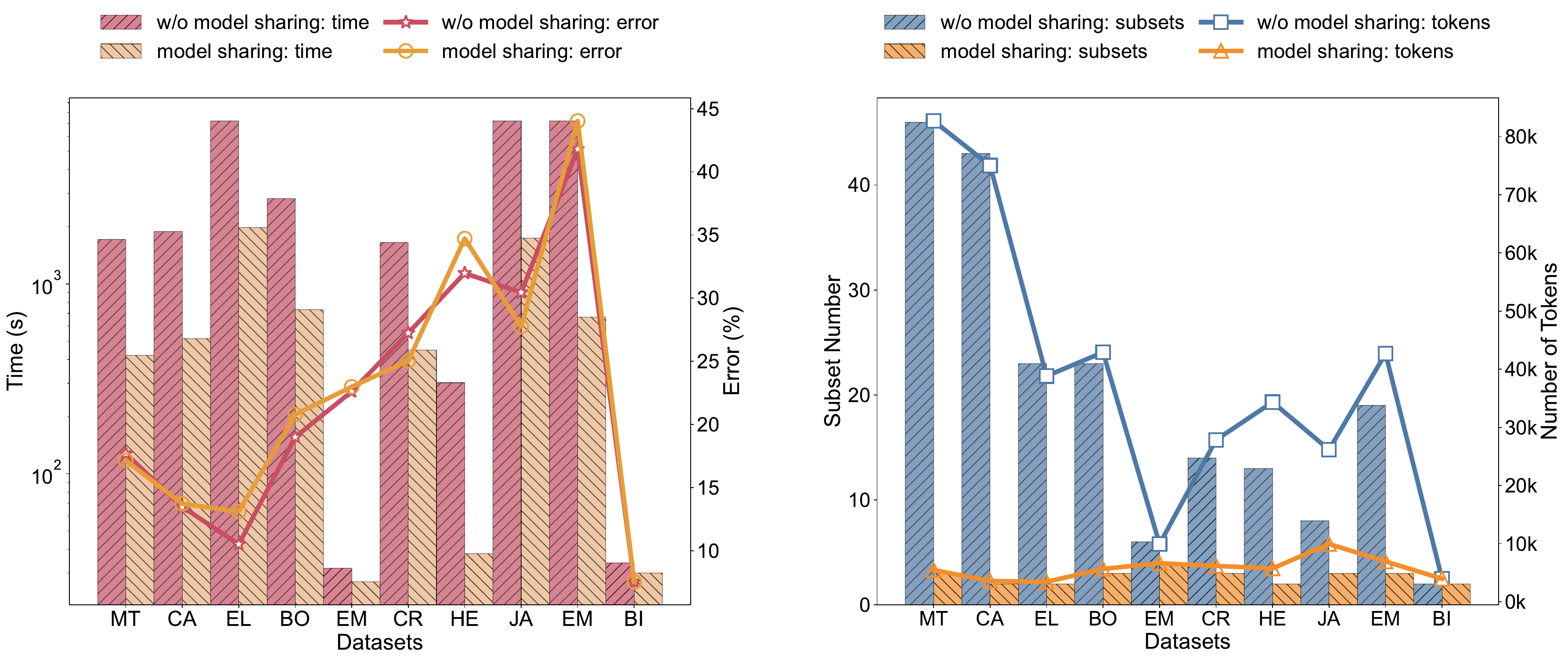}%
    
    % 在图片下方手动添加 (a) 和 (b) 标签
    % \par\medskip % 添加一些垂直间距
    \makebox[0.48\columnwidth][c]{\textbf{(a)} Error-Time analysis}
    \hfill
    \makebox[0.48\columnwidth][c]{\textbf{(b)} Subset-NOT analysis}
    
    \caption{Analyses of the model sharing strategy across all datasets. Our runtime is presented on a logarithmic scale.}
    \label{fig:model_sharing}
\end{figure}

(1) In Figure~\ref{fig:model_sharing}(a), through \textsf{model sharing}, DATE optimized the average runtime by 78.0\% at the cost of only a 2\% increase in average error. 
% As Figure~\ref{fig: cd} shows, there is no statistically significant difference between DATE (w/o m.s., gen.) and DATE (w/o gen.), where $p = 0.441 > \alpha = 0.05$. 
Typically, the approach without \textsf{model sharing} could achieve better performance on some datasets, like HE and BO, with more refined fitting. However, as Algorithm~\ref{alg: partition} proceeds, the DGR becomes excessively long, resulting in overly small partitioned subsets. This led to overfitting on datasets such as CR, JA, and MT. In this case, DATE with \textsf{model sharing} can achieve even better performance.

(2) In Figure~\ref{fig:model_sharing}(b), we observe that \textsf{model sharing} enables DATE to reduce model retraining. As a result. The number of subsets is decreased by an average of 86.3\%, leading to an 85\% reduction in NOT consumption.

\textbf{Effectiveness of DGR-Guided Generation.}
Generation process in DATE consists of two essential strategies: (i) \textsf{decision tree reasoning} (denoted as DT Rea.), and (ii) \textsf{DGR optimization} (donated as DGR Op.). We compare the performance of both strategies in Table~\ref{tab: generation strategy}.

% First, note that the DGRs are sufficiently well-optimized in Algorithm~\ref{alg: partition} even without DGR Optimizer to learn the historical predicate combination from high-performing DGRs. Also, the LLM can improve performance based solely on new paths and validation scores from the explicit decision tree as feedback. However, combining both strategies can lead to even better performance. We believe that decision tree reasoning, which highlights important columns and their threshold values, enables the LLM to understand the data better, resulting in the generation of more contextually relevant and useful DGRs. Moreover, decision trees can be easily represented in natural language using if-else syntax, effectively conveying the information about the data to the LLM.
\begin{table}[tbhp]
\caption{Performance under different generation strategies. We report error rate (\%) for classification tasks.
% The BEST result is shown in \textbf{bold}.
}
\label{tab: generation strategy}
\resizebox{\columnwidth}{!}{%
\begin{tabular}{@{}c|ccccc@{}}
\toprule
                                        &                             &                                                                                    &                              &                               &                                                                                  \\
\multirow{-2}{*}{Dataset}               & \multirow{-2}{*}{Real Data} & \multirow{-2}{*}{\begin{tabular}[c]{@{}c@{}}w/o DTRea.\\ w/o DGROpt.\end{tabular}} & \multirow{-2}{*}{w/o DTRea.} & \multirow{-2}{*}{w/o DGROpt.} & \multirow{-2}{*}{\begin{tabular}[c]{@{}c@{}}w/ DTRea.\\ w/ DGROpt.\end{tabular}} \\ \midrule
BM                                      & 28.99                       & 25.23                                                                              & 24.80                        & 25.55                         & \textbf{20.86}                                                                   \\
EL                                      & 13.44                       & 13.89                                                                              & \textbf{12.78}               & 13.68                         & \textbf{12.78}                                                                   \\
MT                                      & 21.11                       & 17.98                                                                              & 17.69                        & 17.44                         & \textbf{16.58}                                                                   \\
EM                                      & 45.93                       & 44.37                                                                              & 44.29                        & 43.88                         & \textbf{43.10}                                                                   \\
HE & 36.65                       & 36.71                                                                              & 35.81                        & 35.95                         & \textbf{34.50}                                                                   \\
CR                                      & 30.90                       & 29.49                                                                              & \textbf{24.79}               & 24.93                         & \textbf{24.79}                                                                   \\
CA                                      & 15.60                       & 14.38                                                                              & 13.52                        & 13.02                         & \textbf{12.92}                                                                   \\
JA                                      & 30.16                       & 28.44                                                                              & 27.35                        & 26.95                         & \textbf{26.84}                                                                   \\ \midrule
AVG                                     & 27.85                       & 26.31                                                                              & 25.13                        & 25.17                         & \textbf{24.04}                                                                   \\ \bottomrule
\end{tabular}%
}
\end{table}

We observe that the DATE with both strategies consistently achieves the best performance across all classification datasets. 
% outperforming DT Rea. or DGR Op. alone. 
It is worth noticing that \uline{DATE (w/o DT Rea., DGR Op.) still outperforms real data}. LLMs can improve performance based solely on new paths and validation scores from the explicit decision tree as feedback. However, combining both strategies can lead to even better performance. We believe that \textsf{decision tree reasoning} highlights historically effective predicates and steers LLMs to generate more and useful DGRs.
% enabling LLMs to understand the data better, and resulting in the generation of more contextually relevant and useful DGRs.
% Moreover, decision trees can be easily represented in natural language using if-else syntax, effectively conveying the information about the data to LLMs.

\begin{figure}[htbp]
    \centering
    \includegraphics[width=\columnwidth, trim=0 1cm 0 1cm, clip]{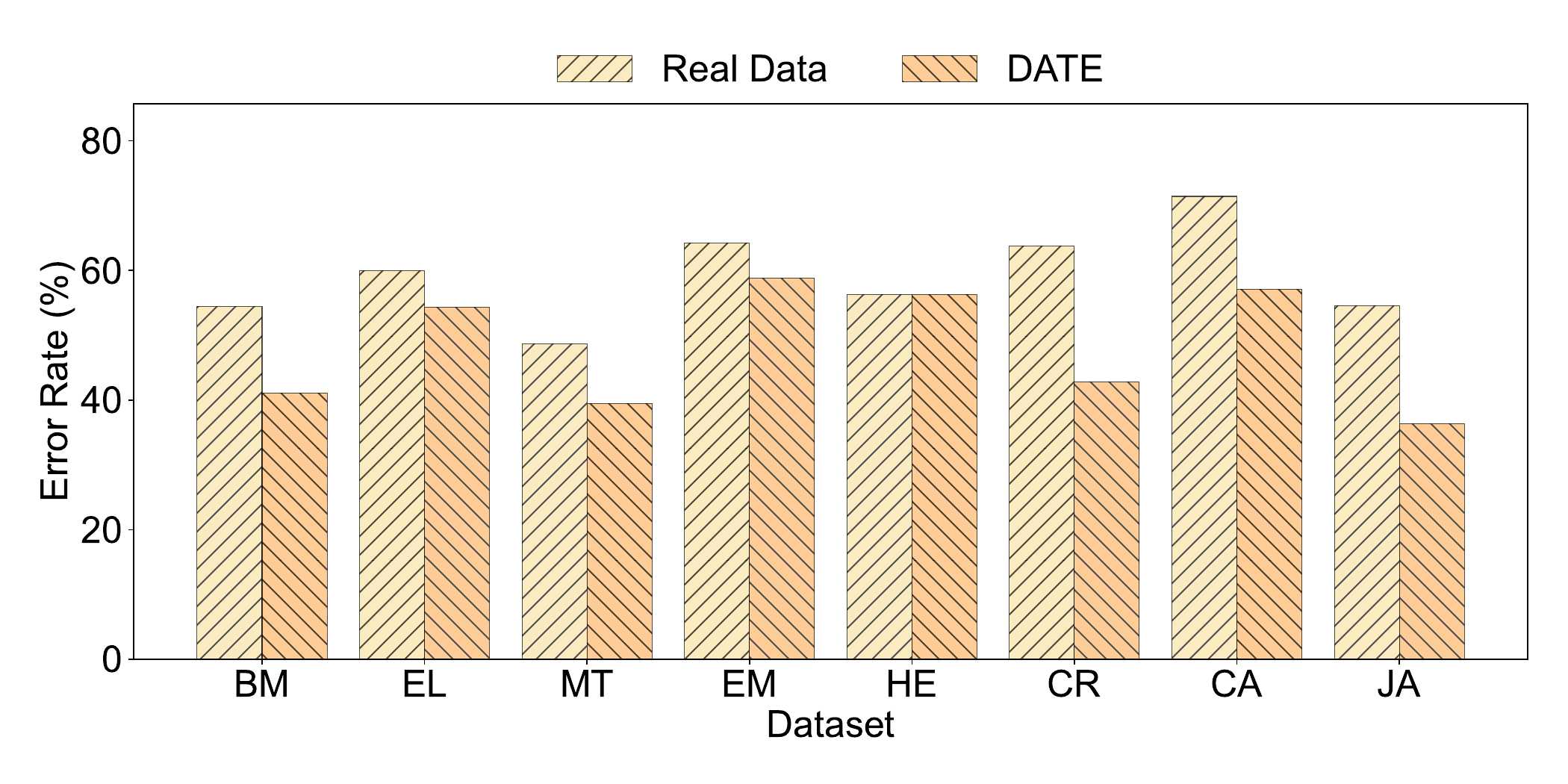}
    \caption{DPO experiments using Q\&A pairs constructed from classification datasets. We report the error rate of the LLM's responses.}
    \label{fig: DPO}
\end{figure}

\textbf{Can DATE Benefit LLM Reasoning?}
While previous research has primarily focused on improving accuracy, we attempt to prove the advancement of our DGR-Guided Generation by answering a truly important question: \uline{To what extent does the DATE-generated data enhance LLM reasoning?} In our generation Algorithm~\ref{alg: generation}, the key idea is to refine DGRs, allowing the LLM to learn more distribution-specific information during inference for generating more diverse data rather than relying on prior knowledge. To evaluate how well the LLM learns distribution preferences in DATE, we compared the performance of Direct Preference Optimization (DPO)~\cite{DPO} with and without DATE's generated data. The details of the DPO experiments can be found in our open-sourced code. The comparison results are shown in Figure~\ref{fig: DPO}.

% 生成增强数据通过多重机制促进DPO微调效果的提升。首先，增强数据有效扩展了训练样本的覆盖范围和多样性，使模型能够学习到更完整的数据分布特征，显著提升了模型的泛化能力。其次，在增强数据中引入规则路径信息，为偏好学习提供了明确的结构化引导，使模型不仅能区分回答的优劣，更能理解优质回答背后的推理逻辑和决策依据。
We observe that DATE significantly improves the accuracy of DPO across all classification tasks by an average of 10.88\%. It demonstrates that the DATE-generated data effectively expands the diversity of training samples, enabling LLMs to learn a more comprehensive set of data distribution features and enhancing their generalization capability. Also, the integration of DGRs provides structural guidance for preference learning. DGR encourages the LLM to focus on how key features influence the quality of good answers, but also to comprehend various distributions, allowing LLMs to quickly adapt to shifts in data distribution in the DPO algorithm.

\textbf{Effectiveness of MDS.}
In this section, we compare the efficiency of our MAB-based generated data sampling algorithm, MDS, with commonly used greedy selection algorithms: Forward Greedy Searching (FGS) and Backward Greedy Search (BGS). We also compare MDS with state-of-the-art data selection methods Quad~\cite{Quad} and Perplexity Correlations (PC)~\cite{Perplexity}, to prove the effectiveness of MDS in selecting quality-diversity balanced data. 

As shown in Table~\ref{tab: MDS}, MDS achieves the best performance across most datasets. For greedy-based methods, they always select data with the best validation score. However, as established by Theorem~\ref{greedy}, the greedy selection property does not hold for heterogeneous data selection, leading to suboptimal results. BGS prioritizes later-generated and multi-round optimized results, achieving the second-best performance. With the novelly proposed DGR-based data diversity, MDS can better capture the impact of different distributions on diversity.
% and thus significantly outperforms all other approaches.

\begin{table}[htbp]
\vspace{-1em}
\caption{Performance under different data selection methods.
% We report \textbf{error rate} for \colorbox[HTML]{CBCEFB}{classification} tasks and \textbf{MSE} for \colorbox[HTML]{FFCCC9}{regression} tasks. THE BEST error in each row is shown in \textbf{bold}, and THE SECOND-BEST error is shown in \underline{underline}. 
}
\label{tab: MDS}
\resizebox{\columnwidth}{!}{%
\begin{tabular}{@{}cc|cc|ccc|c@{}}
\toprule
\multicolumn{2}{c|}{Original Data}                          & \multicolumn{2}{c|}{Data Selection} & \multicolumn{3}{c|}{Greedy-based}          & Ours           \\ \midrule
\multicolumn{1}{c|}{Dataset}                    & Real Data & Quad            & PC                & FGS         & BGS            & Top5        & MDS            \\ \midrule
\multicolumn{1}{c|}{\cellcolor[HTML]{CBCEFB}BM} & 28.99     & 28.02           & 26.20             & {\ul 23.77} & 24.82          & 24.47       & \textbf{20.86} \\
\multicolumn{1}{c|}{\cellcolor[HTML]{CBCEFB}EL} & 13.44     & 13.23           & 13.20             & 13.15       & {\ul 13.06}    & 13.07       & \textbf{12.78} \\
\multicolumn{1}{c|}{\cellcolor[HTML]{CBCEFB}MT} & 21.11     & 22.60           & 21.50             & 17.27       & {\ul 16.79}    & 17.03       & \textbf{16.58} \\
\multicolumn{1}{c|}{\cellcolor[HTML]{CBCEFB}EM} & 45.93     & 44.89           & \textbf{41.80}    & 44.03       & 43.96          & 43.70       & {\ul 43.10}    \\
\multicolumn{1}{c|}{\cellcolor[HTML]{CBCEFB}HE} & 36.65     & 35.85           & 35.90             & 34.72       & \textbf{33.84} & 35.16       & {\ul 34.50}    \\
\multicolumn{1}{c|}{\cellcolor[HTML]{CBCEFB}CR} & 30.90     & 30.57           & {\ul 29.30}       & 30.96       & 30.62          & 30.67       & \textbf{24.79} \\
\multicolumn{1}{c|}{\cellcolor[HTML]{CBCEFB}CA} & 15.60     & {\ul 13.64}     & 14.20             & 14.22       & 13.79          & 13.79       & \textbf{12.92} \\
\multicolumn{1}{c|}{\cellcolor[HTML]{CBCEFB}JA} & 30.16     & 31.04           & 30.00             & 28.49       & {\ul 28.16}    & 28.58       & \textbf{26.84} \\ \midrule
\multicolumn{1}{c|}{CLS-AVG}                    & 27.85     & 27.48           & 26.51             & 25.83       & {\ul 25.63}    & 25.81       & \textbf{24.04} \\ \midrule
\multicolumn{1}{c|}{\cellcolor[HTML]{FFCCC9}BI} & 31.24     & 26.47           & 20.24             & {\ul 8.47}  & 8.52           & 8.47        & \textbf{7.52}  \\
\multicolumn{1}{c|}{\cellcolor[HTML]{FFCCC9}BO} & 22.27     & 32.96           & 35.85             & 22.07       & {\ul 21.51}    & {\ul 21.51} & \textbf{11.79} \\ \midrule
\multicolumn{1}{c|}{REG-AVG}                    & 26.76     & 29.72           & 28.05             & 15.27       & 15.02          & {\ul 14.99} & \textbf{9.66}  \\ \bottomrule
\end{tabular}%
}
\end{table}

\textbf{Takeaway:}
The \textsf{model sharing} strategy significantly reduces the time required for both subset partitioning and data generation. In real-world scenarios, the MDS algorithm proves to be a more practical solution for data selection compared to Quad and Perplexity Correlations, as it effectively balances the critical trade-off between data quality and diversity. Together with DGR-guided generation, these components enable DATE to synthesize targeted, high-quality data efficiently.

\subsection{Sensitivity Analysis}\label{sec: Sensitivity Analysis}
In this section, we conduct the sensitivity analysis on parameters and discuss how to set them in DATE. 

\begin{figure}[hbtp]
\vspace{-1em}
    \centering
    \includegraphics[width=\columnwidth]{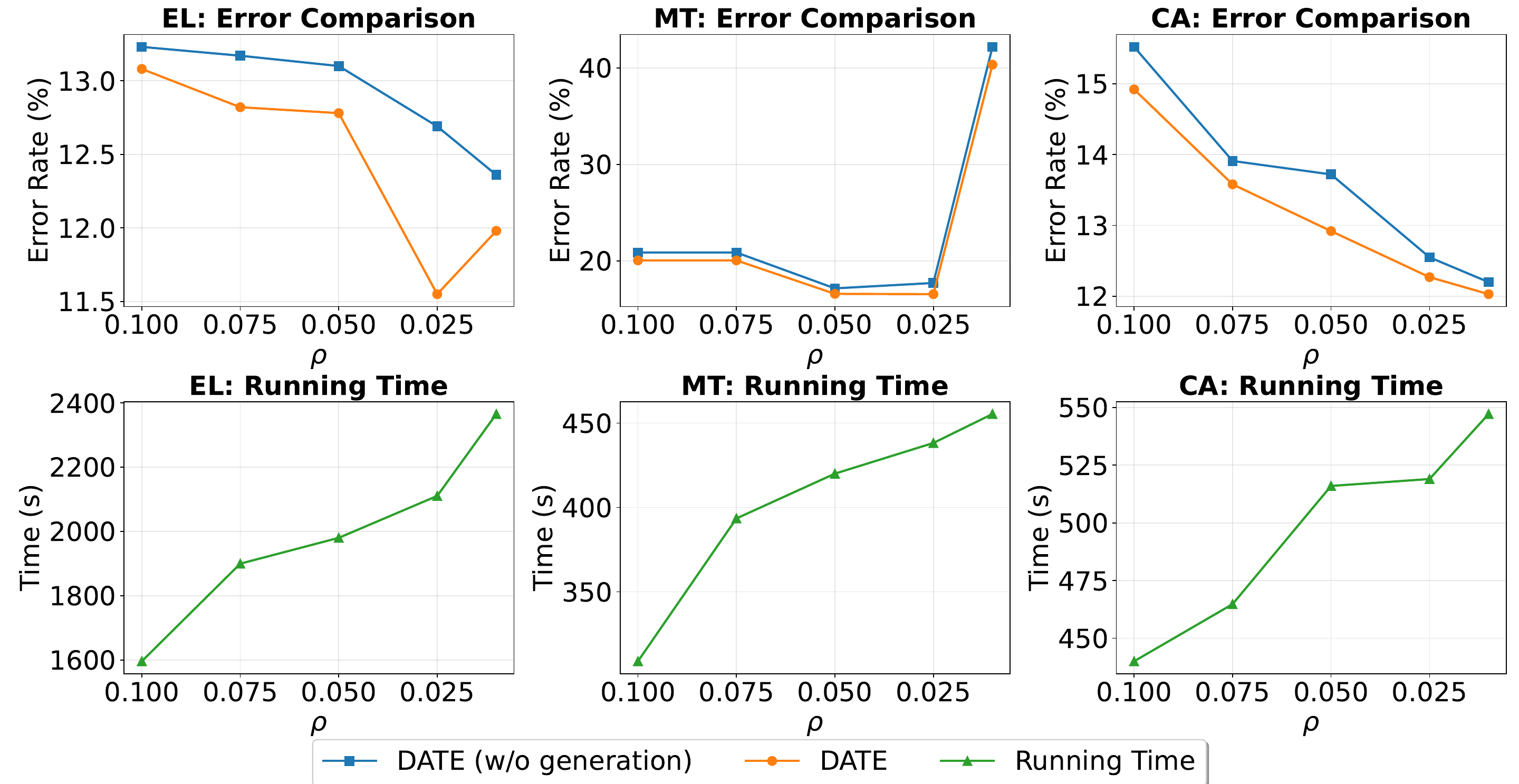}
    \vspace{-1em}
    \caption{The impact of $\rho$ on end-to-end error rate and running time across EL, MT, CA datasets. The horizontal axis is shown in descending order of $\rho$.}
    \label{fig: rou}
\end{figure}

\textbf{Impact of Error Threshold $\rho$.}
% We evaluate the performance with various error thresholds $\rho$ over four datasets. 
The parameter $\rho$ relates to the validation of models $m$ in Algorithm~\ref{alg: partition} and also the quality of generated data in Algorithm~\ref{alg: generation}. Among six figures in Figure~\ref{fig: rou}, we vary the parameter $\rho$, and compare the error rate DATE with or without the generation over three datasets. We also report the running time for Algorithm~\ref{alg: partition}.

% When the parameter $\rho$ is set too lenient, the Algorithm~\ref{alg: partition} runs quickly but fails to partition any subset. 
In Figure~\ref{fig: rou}, as $\rho$ decreases, the error rate gradually diminishes. 
% In this case, we can provide higher-quality DGR-based examples for data generation. 
However, if $\rho$ is set too strictly ($\leq0.025$), while partitioning more detailed diverse distributions, it would lead to increased model retraining and much longer runtime. Also, we observe an abnormal increase in the error rate at $\rho=0.01$ due to model overfitting. To achieve a time-performance tradeoff, the cold-start configuration should set $\rho$ slightly below the original dataset's error rate, thereby ensuring at least one additional subset can be partitioned. In this paper, we set $\rho=0.05$ for classification and $\rho=10$ for regression.

\begin{figure}[htbp]
    \centering
    % 显示整个合并图片
    \includegraphics[width=\columnwidth]{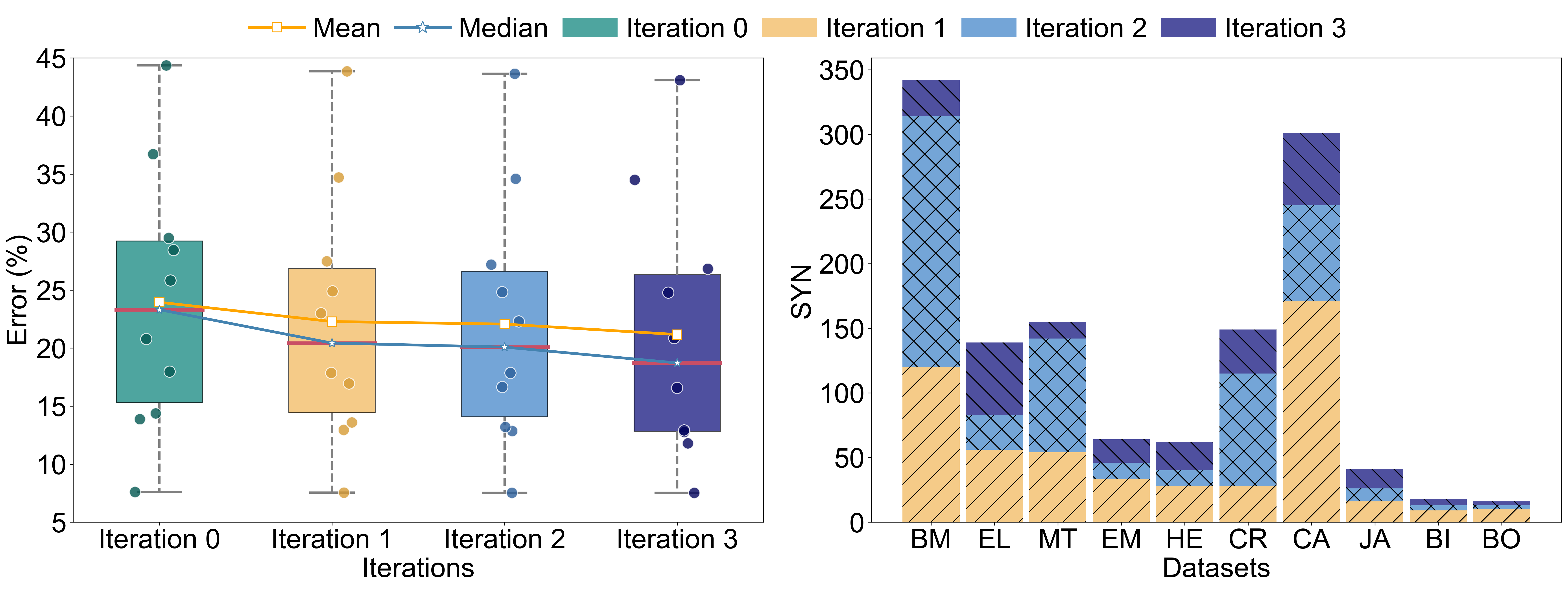}%
    
    % 在图片下方手动添加 (a) 和 (b) 标签
    % \par\medskip % 添加一些垂直间距
    \makebox[0.48\columnwidth][c]{\textbf{(a)} Error analysis}
    \hfill
    \makebox[0.48\columnwidth][c]{\textbf{(b)} SYN analysis}
    \caption{Analyses of $I$ across all datasets.(a) Box plot of error rate across all datasets. The line chart shows the trend of average error as $I$ grows. 
%     % The red line shows the medium error for each iteration.
The scatter points represent the specific error of each dataset. 
    Iteration 0 represents DATE without any generation strategies; (b) Analysis of SYN across all datasets.}
    \label{fig:combined_results}
\end{figure}

\textbf{Impact of Iteration $I$.}
We conducted experiments to evaluate whether increasing the number of iterations $I$ in Algorithm~\ref{alg: generation} helps the LLM refine DGR and generate more useful data. As shown in Figure~\ref{fig:combined_results}, we observed that DATE's performance improved gradually during the initial stages as $I$ increased. However, the synthetic data size decreases as $I$ grows for most datasets. For example, DATE generates only 3 new samples for the Boston dataset in iteration 3. In this case, we limit $I=3$ for all datasets to optimize token usage.

\section{Related Work}\label{sec:Related Work}
Tabular Data Generation (TDG) is a long-standing challenge in data preparation, information retrieval, and natural language processing. We briefly survey the task of TDG, and refer interested readers to~\cite{TDA} for a detailed survey.

\textbf{(1) Non-LLM Tabular Generation Methods.}
 Early approaches of TDG use statistical approaches such as Bayesian networks~\cite{2014PrivBayes} and Fourier decomposition~\cite{Fourier} to model the distribution of the original table and then generate synthetic records by sampling from the distribution. With the rise of generative adversarial networks (GAN)~\cite{table-GAN}, several methods employ GAN to incorporate differential privacy~\cite{PATE-GAN}, preserve functional dependencies between attributes~\cite{ITS-GAN}, or enhance interpretability by modeling feature interactions~\cite{GANBLR}. Specifically, CTGAN~\cite{CTGAN} incorporates techniques for generating discrete variables, but 
 % requires extensive preprocessing to handle mixed data types and 
 struggles when data is scarce. More recently, diffusion models have been applied to improve generation quality and diversity~\cite{STaSy, CoDi, RelDDPM, GOGGLE}. TABSYN~\cite{TABSYN} builds on the score-based generative model by extending diffusion to the VAE latent space and improving support for mixed-type data. These studies aim to learn original distributions like we do, while our DATE harnesses the heterogeneity of tabular data and supports distribution-specific generation.

 % While these methods differ in architecture, they all share a common limitation:  a strong reliance on high-quality and large-scale training data.

\textbf{(2) LLM-based Tabular Generation Methods.}
 LLMs have recently gained attention for tabular data generation. Unlike non-language models that learn only data distributions, LLM-based methods can leverage rich pretrained knowledge, making them well-suited for structured data tasks. Based on how the LLM is used, these approaches can be broadly categorized into two groups: finetuning methods and prompt-based methods. Fine-tuning methods like GReaT~\cite{GReaT} help LLMs model feature dependencies and domain constraints for more coherent outputs by updating LLM weights. HARMONIC~\cite{HARMONIC} fine-tunes LLMs using instructions derived from nearest-neighbor relationships, encouraging LLMs to learn inter-row patterns rather than memorizing records. Although fine-tuning enables high-fidelity generation with sufficient data, it is prone to severe overfitting in low-data regimes~\cite{CLLM}. Prompt-based methods such as EPIC~\cite{EPIC} and CLLM~\cite{CLLM} leverage the in-context learning ability of LLMs, enabling them to generate tabular data by conditioning on a few labeled examples embedded directly in the prompt to guide the generation process without modifying model parameters. These methods struggle to capture logical dependencies and often suffer from low-quality prompts, while our DATE framework improves the quality of designed prompts and can generate diverse data, outperforming existing LLM-based methods. 

\section{Conclusion}
In this paper, we propose DATE, a LLM-based framework for heterogeneous tabular data generation. We derived a data partitioning algorithm to discovery Distribution-Guided Rules (DGR) for constructing LLM-friendly prompts. To enhance LLM adaptation to the target dataset, we design generation based on decision tree reasoning and DGR optimization. For iteratively generated data, we prove that the selection process is non-greedy with respect to heterogeneity and propose a novel error-guaranteed sampling method. Our experiments show that DATE outperforms SOTA GAN-based and LLM-based solutions, achieving the most compact yet precise heterogeneous data generation. Our Direct Preference Optimization experiment further confirms that DATE-generated data can effectively steer LLM reasoning on the target dataset.

 %\clearpage
\bibliographystyle{IEEEtran}
\bibliography{simple}

\begin{IEEEbiography}
[{\includegraphics[width=1in,height=1.25in,clip,keepaspectratio]{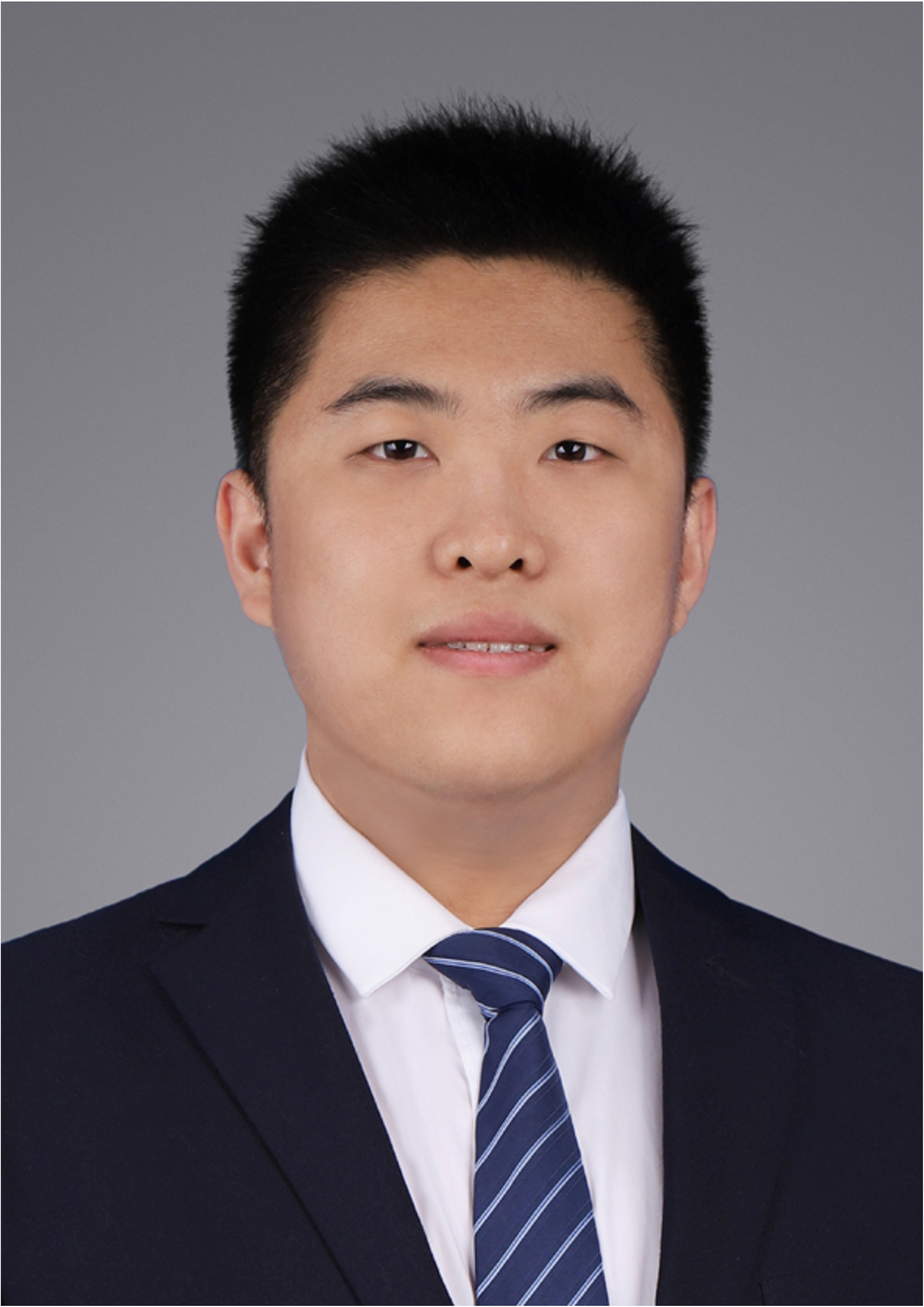}}]
{Yafeng Tang} is a Ph.D. student in computer science at the School of Computer Science and Technology, Harbin Institute of Technology, China, under the direction of Prof. Hongzhi Wang. He received his bachelor's degree from Harbin Institute of Technology. His primary research interests include data preparation, data quality management, and automatic machine learning. He is an IEEE Student Member.
\end{IEEEbiography}

\begin{IEEEbiography}
[{\includegraphics[width=1in,height=1.25in,clip,keepaspectratio]{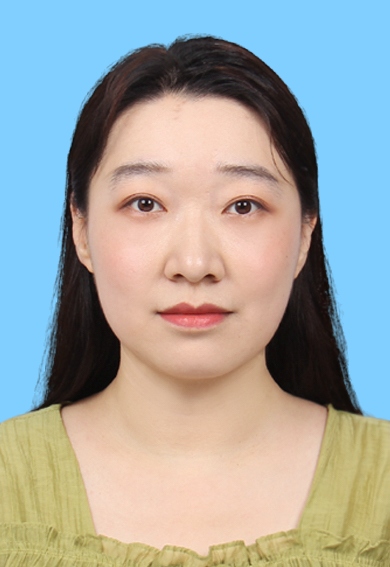}}]
{Xiaoou Ding}, Associate Professor, Doctoral Supervisor in the Department of Computing at Harbin Institute of Technology. She received both her Bachelor's and Doctoral degrees from HIT. She currently serves as an Executive Member of the Database Special Committee of the China Computer Federation (CCF), an Executive Member of the CCF Data Governance Development Committee. Her research interests include big data governance, data quality management, and high-quality data construction for machine learning. She has presided over numerous research projects, including the National Natural Science Foundation of China (NSFC) General Program, NSFC Youth Program, sub-project of the National Key R\&D Program, Heilongjiang Provincial Natural Science Foundation Excellent Youth Program, CCF-Huawei Poplar Forest Database Special Project, and China Postdoctoral Science Foundation Project. As the first/corresponding author, she has received the First Prize of CCF Computer Application Innovation Product Award (2024), ACM China Harbin Chapter Rising Star Award (2023), Leader 5000 - Top Paper Award of China's Elite Scientific Journals (2024, 2022), and High-Impact Paper Awards from the Journal of Software (2022, 2023).
\end{IEEEbiography}

\begin{IEEEbiography}
[{\includegraphics[width=1in,height=1.25in,clip,keepaspectratio]{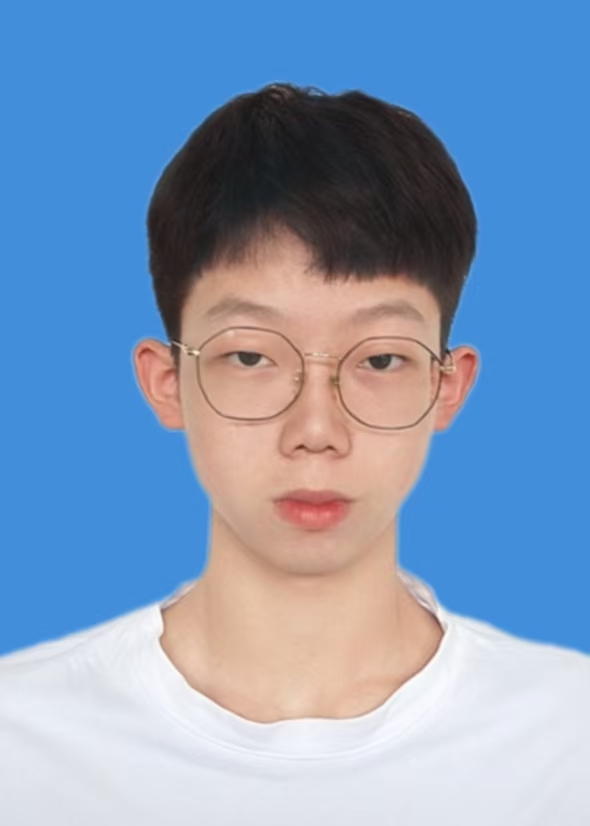}}]
{Jianzhuo Du} is an undergraduate student in Computer Science at the School of Computer Science and Technology, Harbin Institute of Technology, China, under the supervision of Professor Hongzhi Wang. His research focuses on data preparation for large language models.
\end{IEEEbiography}

\begin{IEEEbiography}
[{\includegraphics[width=1in,height=1.25in,clip,keepaspectratio]{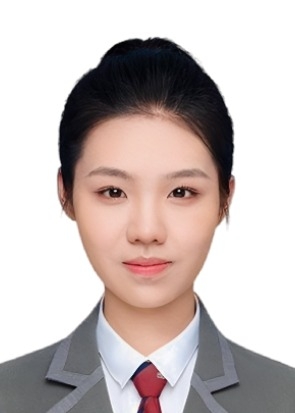}}]
{Zishuo Yan} is an undergraduate student in Computer Science at the School of Computer Science and Technology, Harbin Institute of Technology, China, under the supervision of Professor Hongzhi Wang. Her research focuses on data quality management.
\end{IEEEbiography}

\begin{IEEEbiography}
[{\includegraphics[width=1in,height=1.25in,clip,keepaspectratio]{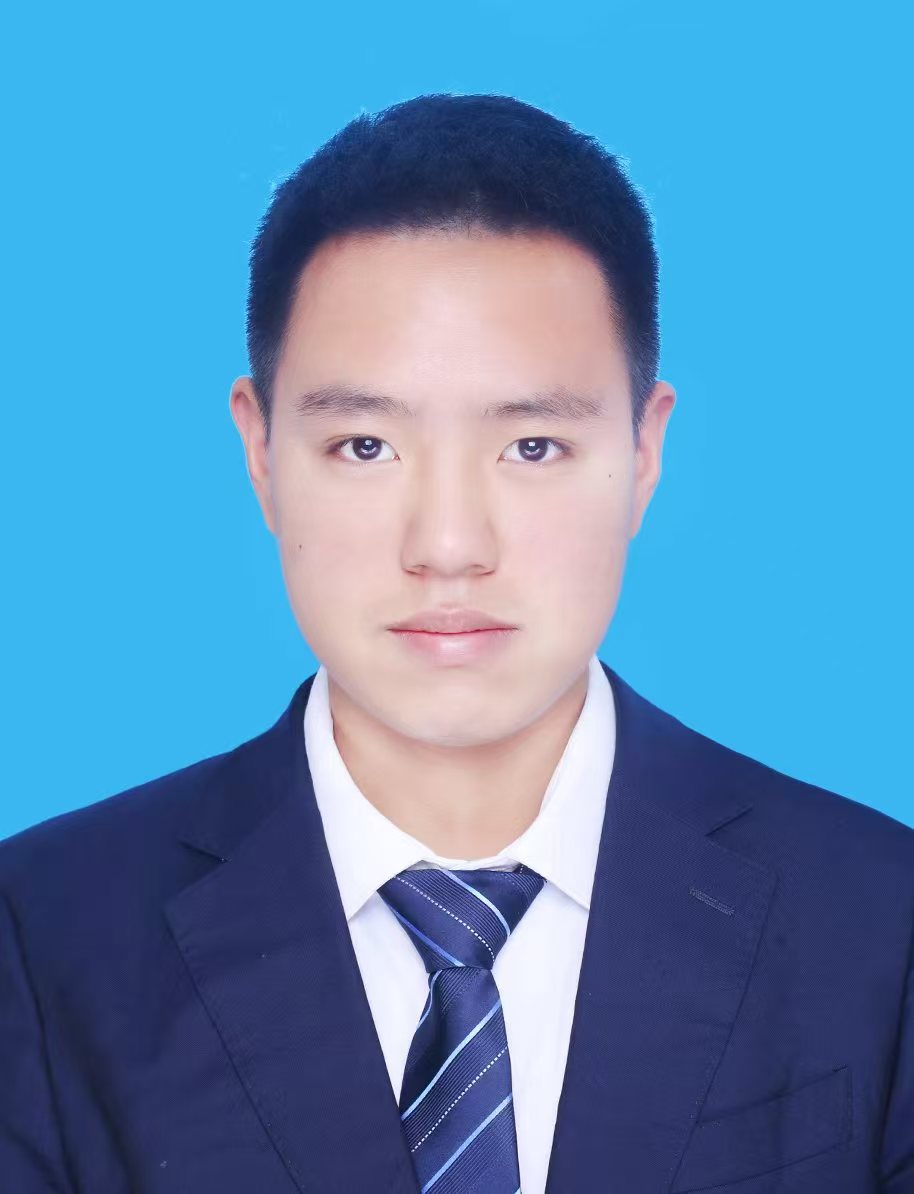}}]
{Zhuang Ma} is an undergraduate student in Computer Science at the School of Computer Science and Technology, Harbin Institute of Technology, China, under the supervision of Professor Hongzhi Wang. His research focuses on data quality management.
\end{IEEEbiography}

\begin{IEEEbiography}
[{\includegraphics[width=1in,height=1.25in,clip,keepaspectratio]{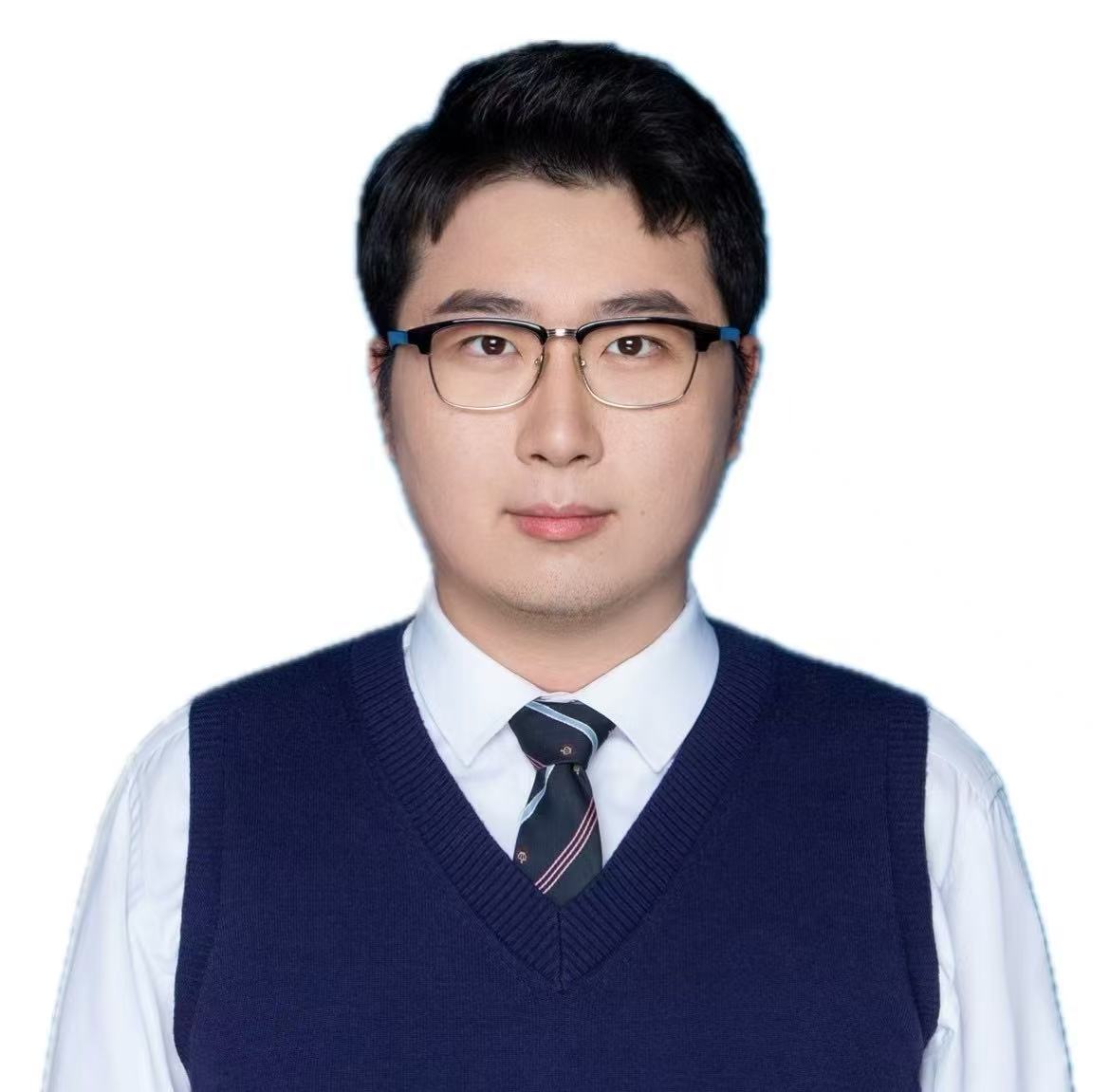}}]
{Zheng Liang} is a Ph.D. student in computer science at the School of Computer Science and Technology, Harbin Institute of Technology, China, under the supervision of Prof. Hongzhi Wang. He received his bachelor’s degree from Harbin Institute of Technology. His research focuses on large language model-based data wrangling, time series data analysis, and automated machine learning.
\end{IEEEbiography}

\begin{IEEEbiography}
[{\includegraphics[width=1in,height=1.25in,clip,keepaspectratio]{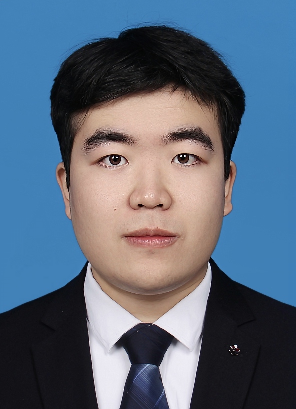}}]
{Zekai Qian} is a Ph.D. student in computer science at the School of Computer Science and Technology, Harbin Institute of Technology, China, under the supervision of Prof. Hongzhi Wang. He received his bachelor’s degrees from Harbin Institute of Technology. His research focuses on data quality management, including high-quality data preparation, model-aware data cleaning, and data-centric AI.
\end{IEEEbiography}

\begin{IEEEbiography}
[{\includegraphics[width=1in,height=1.25in,clip,keepaspectratio]{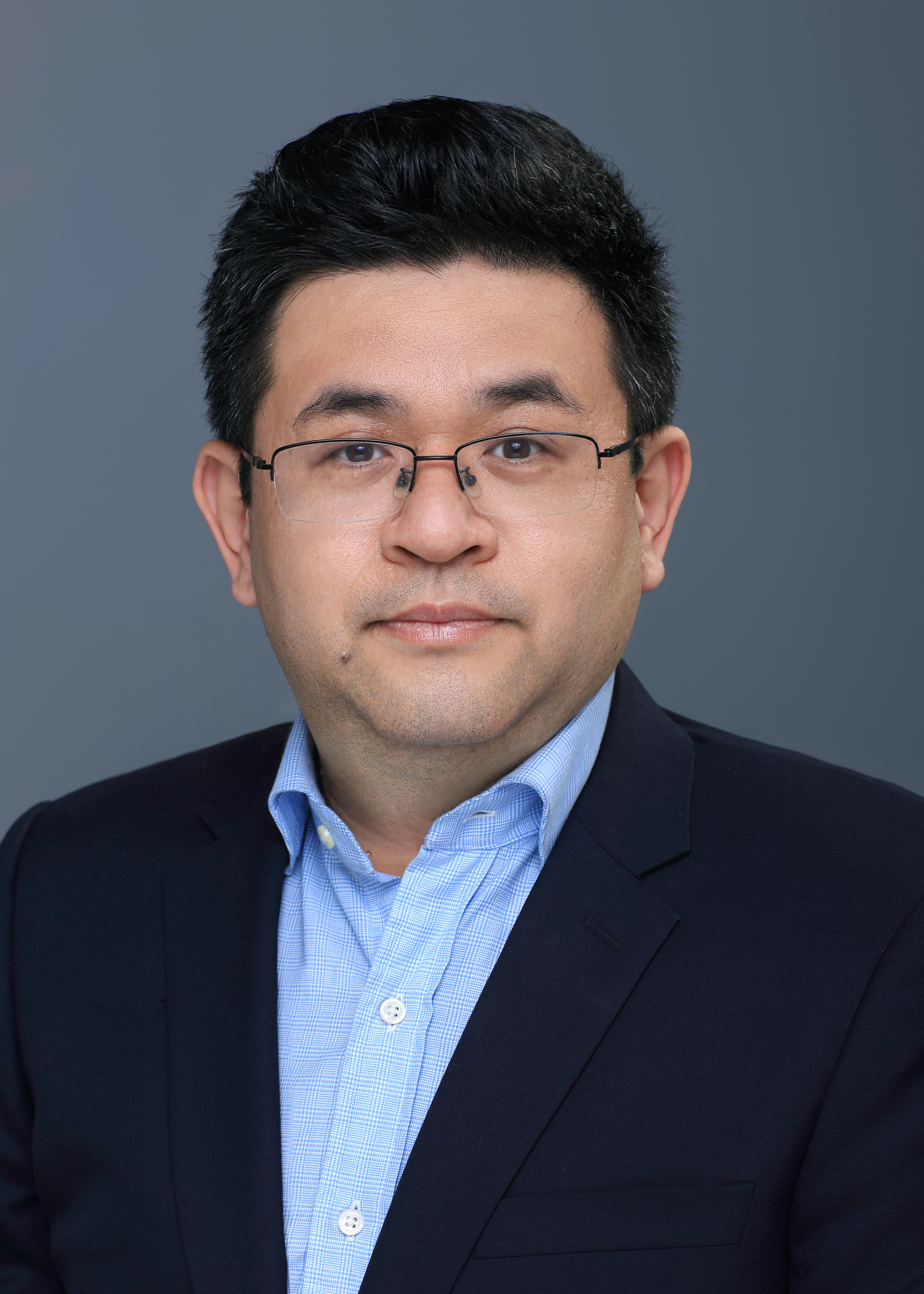}}]
{Hongzhi Wang}, Professor, PHD supervisor, the head of massive data computing center, the secretary general of ACM SIGMOD China, outstanding CCF member, a standing committee member CCF databases and a member of CCF big data committee. Research Fields include big data management and analysis, database systems, knowledge engineering and data quality. He was “starring track” visiting professor at MSRA and postdoctoral fellow at University of California, Irvine. Prof. Wang has been PI for more than 10 national or international projects including NSFC key project, NSFC projects and National Technical support project, and co-PI for more than 10 national projects include 973 project, 863 project and NSFC key projects. He also serves as a member of ACM Data Science Task Force. He has won First natural science prize of Heilongjiang Province, MOE technological First award, Microsoft Fellowship, IBM PHD Fellowship and Chinese excellent database engineer. His publications include over 200 papers in the journals and conferences such as VLDB Journal, IEEE TKDE, VLDB, SIGMOD, ICDE and SIGIR, 6 books and 6 book chapters. His PHD thesis was elected to be outstanding PHD dissertation of CCF and Harbin Institute of Technology. His papers were cited more than 6500 times. His personal website is http://homepage.hit.edu.cn/wang.
\end{IEEEbiography}
\newpage

\vfill

\end{document}